\documentclass[10pt,twocolumn,letterpaper]{article}

\usepackage[pagenumbers]{wacv} 

\definecolor{wacvblue}{rgb}{0.21,0.49,0.74}
\usepackage[pagebackref,breaklinks,colorlinks,allcolors=wacvblue]{hyperref}
\usepackage{stmaryrd}
\usepackage[accsupp]{axessibility}
\def\wacvPaperID{277} 
\def\confName{WACV}
\def\confYear{2026}

\title{Shift-Equivariant Complex-Valued Convolutional Neural Networks}

\author{\textbf{Quentin Gabot}\textsuperscript{1,3}, \textbf{Teck-Yian Lim}\textsuperscript{4}, \textbf{Jérémy Fix}\textsuperscript{2}, \textbf{Joana Frontera-Pons}\textsuperscript{3},\\\textbf{Chengfang Ren}\textsuperscript{1} and \textbf{Jean-Philippe Ovarlez}\textsuperscript{1,3}\\
\small \textsuperscript{1} SONDRA, CentraleSupélec, Université Paris-Saclay, Gif-sur-Yvette, France\\
\small \textsuperscript{2} LORIA, CNRS, CentraleSupélec, Université Paris-Saclay, France\\
\small \textsuperscript{3} DEMR, ONERA, Université Paris-Saclay, Palaiseau, France\\
\small \textsuperscript{4} DSO National Laboratories, Singapore\\
}

\begin{document}
\maketitle
\begin{abstract}
Convolutional neural networks have shown remarkable performance in recent years on various computer vision problems. However, the traditional convolutional neural network architecture lacks a critical property: shift equivariance and invariance, broken by downsampling and upsampling operations. Although data augmentation techniques can help the model learn the latter property empirically, a consistent and systematic way to achieve this goal is by designing downsampling and upsampling layers that theoretically guarantee these properties by construction. Adaptive Polyphase Sampling (APS) introduced the cornerstone for shift invariance, later extended to shift equivariance with Learnable Polyphase up/downsampling (LPS) applied to real-valued neural networks. In this paper, we extend the work on LPS
to complex-valued neural networks both from a theoretical perspective and with a novel building block of a projection layer from $\mathbb{C}$ to $\mathbb{R}$ before the Gumbel Softmax. We finally evaluate this extension on several computer vision problems, specifically for either the invariance property in classification tasks or the equivariance property in both reconstruction and semantic segmentation problems, using polarimetric Synthetic Aperture Radar images.
\end{abstract}

\section{Introduction} \label{sec:intro}
Since their introduction by LeCun et al.~\cite{lecun1995convolutional}, Convolutional Neural Networks (CNNs) have remained the standard neural network architecture for dealing with spatially ordered data, such as images. 
Intuitively, many computer vision tasks rely on patterns that can appear anywhere in the image. Convolutional layers in CNNs introduce this specific bias into neural networks, as it is the `sliding' of a kernel over the entire image. Convolutional layers are also a special case of dense linear layers, $y=Wx+b$, whereby $W$ is a Toeplitz matrix instead of any matrix. 
\\
However, modern CNN architectures include downsampling and upsampling layers that break equivariance. It has been empirically shown that resultant networks are highly sensitive to translations in the input \cite{zhang2019making}. 
To maintain the desirable shift equivariant property, recent works have explored provably equivariant subsampling layers \cite{chaman2021truly, rojas2022learnable}.
This search for invariance and equivariance to known symmetries in input data falls into a more general research topic: geometric deep learning \cite{bronstein2017geometric}. 
This branch of deep learning explores methods for accounting for the fact that the use of geometric priors can influence the design of neural networks.
\\
While well-developed for real-valued data, this topic is far less studied in the field of complex-valued neural networks, where both the input data and the neural network parameters are composed of complex numbers. \cite{trabelsi2017deep, hirose_complex-valued_2009}.
Imaging modalities commonly encountered in remote sensing (e.g., Synthetic Aperture Radar (SAR) \cite{Pottier09}) and medical imaging (e.g., Magnetic Resonance Imaging (MRI) \cite{cole2021analysis}) often consist of naturally complex-valued images.
The typical approach in dealing with such data is often only utilizing the magnitude of the complex numbers, even though rich information can be embedded in the phase \cite{barrachina2023comparison, cvwgan, gabotpreserving}. 
Complex-valued deep learning is an active research topic, as it has the potential to discover and exploit patterns that are either overlooked or yet to be discovered.
Several foundational works had already been published \cite{ConvModelICCV, geuchen2023optimal, wu2023complex}.
\\
\\
\textbf{The key contributions of this work are:}
\begin{itemize}
\item The extension of provable shift-equivariance to the complex domain,
\item A learnable complex-to-real projection,
\item The empirical results demonstrating the benefits of a fully complex neural network for naturally complex data on three computer vision tasks (classification, reconstruction, segmentation).
\end{itemize}
In Section \ref{sec:related}, we present the existing works related to key concepts like shift-invariant/equivariant CNNs and CVNNs. In Section \ref{sec:shift-inv/eq}, we extend the theory of shift-invariant/equivariant convolutional neural networks to the complex domain by providing a theoretical explanation of their design. Section \ref{sec:experiments} describes an extensive set of experiments to demonstrate the impact of the shift-invariance/equivariance property on CVNNs. Finally, Section \ref{sec:discussions} discusses some perspectives in remote sensing \cite{mian2019design}, medical imaging \cite{dedmari2018complex}, and earth monitoring \cite{patruno2013polarimetric}. 
\\
\\
\textbf{Notations}: Italic type indicates a scalar quantity $x$, lower case boldface indicates a vector quantity $\mathbf v$, and upper case boldface indicates a matrix or tensor $\mathbf A$. $\Re{(z)}$ and $\Im{(z)}$ respectively denote the real and imaginary parts of complex number $z\in \mathbb{C}$. $\lfloor x \rfloor$ is the mathematical floor function. The modulo operation is defined as $\mathrm{mod}$. For any vector $\mathbf{v}$, $\mathbf{v}^T$ denotes the transpose operator and $\mathbf{v}[.]$ is the indexing operator on $\mathbf{v}$. $\left\|\mathbf{v}\right\|_2$ stands for the Euclidean norm. The operator $\odot$ is the Hadamard element-wise product between vectors. Finally, the composition operator $\circ$ is defined by $(f \circ g)(x)=f(g(x))$.
\section{Related work} \label{sec:related}
\textbf{Shift-invariant/equivariant convolutional neural networks.} As discussed in Section \ref{sec:intro}, while convolution operations (using a stride of $1$) are indeed shift-equivariant, modern CNNs are not. As explained by Azulay et al. \cite{azulay2019deep}, the traditional downsampling scheme, namely pooling layers, involving a stride strictly larger than $1$, is responsible for breaking this property. We present such a case in Appendix \ref{app:shift}.
\\
Some methods have tried to mitigate the negative effect of the subsampling scheme, for instance, by using anti-aliasing, like in the Low-Pass Filtering (LPF) method proposed by Zhang et al. \cite{zhang2019making, zou2023delving}. However, as demonstrated by Chaman et al. \cite{chaman2021truly}, LPF reduces the effect of downsampling layers but does not solve the shift-equivariance problem.
\\
 The first consistent shift-equivariant method was introduced by Chaman et al. with the Adaptive Polyphase Sampling (APS) method \cite{chaman2021truly}. APS relies on a shift-invariant sub-sampling scheme by selecting the highest $\ell_p$-norm among the polyphase components, as formally defined in Section \ref{sec:lps}. 
\\
Lastly, Lim et al. \cite{rojas2022learnable} further elaborate on this concept and propose a generalization of APS with the Learnable Polyphase Sampling (LPS) method. Indeed, while APS relies on the $\ell_p$ norm to select the polyphase components, LPS introduces a pair of trainable \textless shift-invariant/equivariant\textgreater \textless down/up-sampling\textgreater layers. This property enables an elegant method for selecting a downsampling scheme trained in conjunction with the rest of the network.
Shift-invariance is desirable for classification problems, while shift-equivariance is necessary for tasks such as segmentation and reconstruction. In the latter, a translation of the input should result in the same output translation.
\\
Shift-invariant/equivariant methods have also been extended to vision transformers (ViT) \cite{rojas2024making}. Finally, shift-equivariance is one special case of a broader research topic, group-equivariance, discussed in \cite{cohen2016group, weiler2019general}.
\\
\\
\textbf{Complex-valued neural networks.} Complex-valued neural networks (CVNNs) are deep neural networks whose operations and inputs are defined in the complex domain \cite{trabelsi2017deep, Barrachina2021, hirose_complex-valued_2009}. Despite the novelty of complex-valued neural network theory, several critical questions regarding CVNNs are still open, one being their suitable representation.
\\
Based on the isomorphism between $\mathbb{C}$ and $\mathbb{R}^2$, CVNNs can be either represented as~\cite{lee2022complex}:
\begin{itemize}
    \item A dual-RVNN is a neural network with representations in $\mathbb{R}$, where real and imaginary parts are stacked at the input. Operations are then performed in $\mathbb{R}$,
    \item A split-CVNN processes representations in $\mathbb{R}^2$ with operations in $\mathbb{R}$. Compared to a dual-RVNN, we keep the relationship between real and imaginary components, although the weights are also real-valued.
    \item A full-CVNN contains both representations and operations in $\mathbb{C}$.
\end{itemize}
However, these neural networks are not equivalent: dual-RVNNs lack any representation of the complex domain. Meanwhile, Hirose et al. \cite{hirose2012generalization} argue that fully-CVNNs are better representation choices for CVNNs. Finally, Wu et al. \cite{wu2024complex} recently proved that complex-valued neurons can learn more than real-valued neurons, reigniting discussions surrounding this question.
\\
The core components of neural networks have been largely extended to the complex domain. For instance, Li et al. \cite{li2008complex} proposed using Wirtinger calculus to define complex-valued backpropagation properly, and more recently, Trabelsi et al. \cite{trabelsi2017deep} extended the batch normalization layer to the complex domain.
\\
In this paper, we extend the LPS approach to CVNNs, specifically fully-CVNNs, and compare it to RVNNs, namely dual-RVNNs. 
\section{Complex-valued shift-invariant/equivariant layers} \label{sec:shift-inv/eq}
As discussed in Section \ref{sec:related}, shift-equivariance is a highly desired property for complex-valued neural networks. While shift-equivariant methods for neural networks have been proposed and tested for the real domain, an extension of these methods to the complex domain has yet to be realized. Among the various algorithms, such as APS \cite{chaman2021truly} and LPS \cite{rojas2022learnable}, LPS is the current state-of-the-art shift-equivariant method for RVNNs. As such, our contributions focus on its extension to the complex domain. We will also introduce a learnable projection function to address some difficulties encountered during the extension of the LPS method to the complex domain.

\subsection{Preliminaries} \label{sec:preli}
In this section, we formally introduce crucial concepts related to shift-invariance/equivariance and CVNNs. Note that every notion discussed in this section is defined by default in the complex domain. For the sake of simplicity, we consider, in the following, 1D signals and downsampling factor $p=2$, although the method is generalizable for higher dimensions and downsampling factors.
\\
\\
\textbf{Shift-equivariance.} Shift-equivariance is a property attributed to functions whose outputs are shifted accordingly to the shift in their inputs. Formally, a function $f: \mathbb{C}^{N} \mapsto \mathbb{C}^{M} $ is $\mathbf{T}_{N},\{\mathbf{T}_{M}, \mathbf{I}\}$-equivariant (i.e. shift-equivariant) iif:
\begin{equation}
    \exists \mathbf{T} \in\{\mathbf{T}_{M}, \mathbf{I}\}, \ \forall \mathbf{z} \in \mathbb{C}^{N},\ f\left(\mathbf{T}_{N} \mathbf{z}\right) = \mathbf{T} f\left(\mathbf{z}\right),
\end{equation}
where $\forall n \in \mathbb{Z}, \, \displaystyle \mathbf{T}_{N}\mathbf{v}[n] \triangleq \mathbf{v}\left[(n+1) \, \mathrm{mod} \ N\right]$
is a circular shift.
\\
\\
\textbf{Shift-invariance.} The shift-invariance property refers to the case where an output is never affected by any shift in the input. Formally, a function $f: \mathbb{C}^{N} \mapsto \mathbb{C}^{M} $ is $\mathbf{T}_{N},\{\mathbf{I}\}$-invariant (i.e. shift-invariant) iif:
\begin{equation}
    f(\mathbf{T}_{N} \mathbf{z}) = f(\mathbf{z})\, ,  \forall \mathbf{z} \in \mathbb{C}^{N}\, .
    \label{eq:invariance}
\end{equation}
We obtain shift-invariance from shift-equivariant functions by applying a permutation-invariant linear form (an operator mapping from $\mathbb{C}^N$ to $\mathbb{C}$ invariant under any permutation of the components of its input). One such example is the sum operation, as we observe that:
\begin{equation}
    \sum_{m} f\left(\mathbf{T}_{N} \mathbf{z}\right)[m] = \sum_{m} \left(\mathbf{T}_{M} f(\mathbf{z})\right)[m]\, ,
    \label{eq:invariant}
\end{equation}
is shift-invariant if $f$ is shift-equivariant. From a more general perspective, based on the above claim, any global pooling operator, i.e., an operator with a global receptive field (max, mean, sum, or product), is a shift-invariant transform.
\\
\\
\textbf{Complex-valued downsampling \& pooling layer.} Unlike the previously introduced global pooling, pooling layers with a local receptive field are not shift-invariant/equivariant. Indeed, commonly used pooling layers are fundamentally a filter (max, mean) with a stride of $1$ followed by a subsampling (using a stride strictly larger than $1$), which is not a shift-invariant/equivariant operator. Indeed, a downsampling-by-$p$ factor operator $\mathbf{D}_p:  \mathbb{C}^{N} \mapsto \mathbb{C}^{ \left\lfloor N/p \right\rfloor}$ is defined as:
\begin{equation}
    \mathbf{D}_p(\mathbf{z})[n]=\mathbf{z}[p \, n] \, , \ \forall n \in \left\llbracket 1, \left\lfloor N/p \right\rfloor \right\rrbracket.
    \label{eq:downsampling}
\end{equation}
Such an operator constantly samples the same indices of its input, regardless of whether a shift occurred.
\\ 
Note that some pooling layers in the complex domain differ from their real-valued counterparts. The max pooling layer cannot be trivially extended to the complex domain as $\mathbb{C}$ is only partially ordered. As such, the returned element is a complex-valued scalar such that the selection rule is ordered, the complex-valued max pooling layer $\mathbf{M}: \mathbb{C}^{N} \mapsto \mathbb{C}$ being defined as:
\begin{equation}
    \mathbf{M}(\mathbf{z})=\mathbf{z}[k] \mbox{ where } k = \arg \max_{i\in \left\llbracket 1,n \right\rrbracket}\left|\mathbf{z} [i]\right| \, ,
\end{equation}
\textbf{Complex-valued activations.}
Several activation functions have been proposed in the literature: Kuroe et al. \cite{kuroe2003activation} proposed to distinguish between them by categorizing activation functions into the following two classes of complex functions $f(\cdot)$:
\begin{equation}
\begin{array}{l}
    \text{Type A}: f(z) = f_{\Re}(\Re(z)) + j\, f_{\Im}(\Im(z))\, ,\\
    \text{Type B}: f(z) = \psi(r)\, e^{j\phi(\theta)}\, ,
    \end{array}
\end{equation}
where $f_{\Re}: \mathbb{R}\mapsto\mathbb{R}$ and $f_{\Im}: \mathbb{R}\mapsto\mathbb{R}$ are nonlinear real-valued functions. $\psi: \mathbb{R}^{+}\mapsto\mathbb{R}^{+}$ and $\phi: \, \mathbb{R}\mapsto\mathbb{R}$ are also nonlinear real functions. We select the $\mathrm{modReLU}(.)$ function \cite{arjovsky2016unitary}, such as given $b\in\mathbb{R}$ a bias parameter of the nonlinearity, $\mathrm{modReLU}(.)$ is defined as :
\begin{align}
    \mathrm{modReLU}(\mathbf{z}) & = \mathrm{ReLU}\left(|\mathbf{z}|+b\right)\, e^{j\theta} \nonumber \\
    &= 
    \begin{cases}
    \left(|\mathbf{z}|+b\right)\displaystyle\frac{\mathbf{z}}{|\mathbf{z}|} & \text{if}\ |\mathbf{z}|+b>0\, , \\
      0 & \text{otherwise}\, .
    \end{cases}
\end{align}
\\
\textbf{Complex differentiability.}
The partially ordered nature of $\mathbb{C}$ is incompatible with minimization and optimization. Thus, loss functions $L$, estimating the correctness of the model prediction, must return a real-valued scalar ($L:\mathbb{C}^{n} \mapsto \mathbb{R}$). One common way to define such functions is to consider only the magnitude of $\mathbf{z}$ or compute the loss function on the real and imaginary parts separately.
\\
Additionally, on the topic of complex differentiability, since the loss $L$ is lower bounded, it is not holomorphic by Liouville's theorem. Therefore, the Wirtinger derivative is used to compute the gradient of loss $L$ with respect to the real and imaginary parts of the neural network weights in the complex backpropagation algorithm \cite{sarroff2015learning}.
\\
\\
\textbf{Complex-valued neural networks.}
In addition to pooling layers, loss functions, and activation functions, further building blocks are required to achieve a complex-valued neural network. Linear and convolutional layers extend naturally to the complex domain if their weights are complex-valued. As explained by Trabelsi et al. \cite{trabelsi2017deep}, complex-valued batch normalization and initialization layers require specific implementations, and we followed their proposal in our experiments.
\subsection{Extension of shift-equivariant methods to the complex domain} \label{sec:lps}
\begin{figure*}
    \centering
    \includegraphics[width=0.9\linewidth]{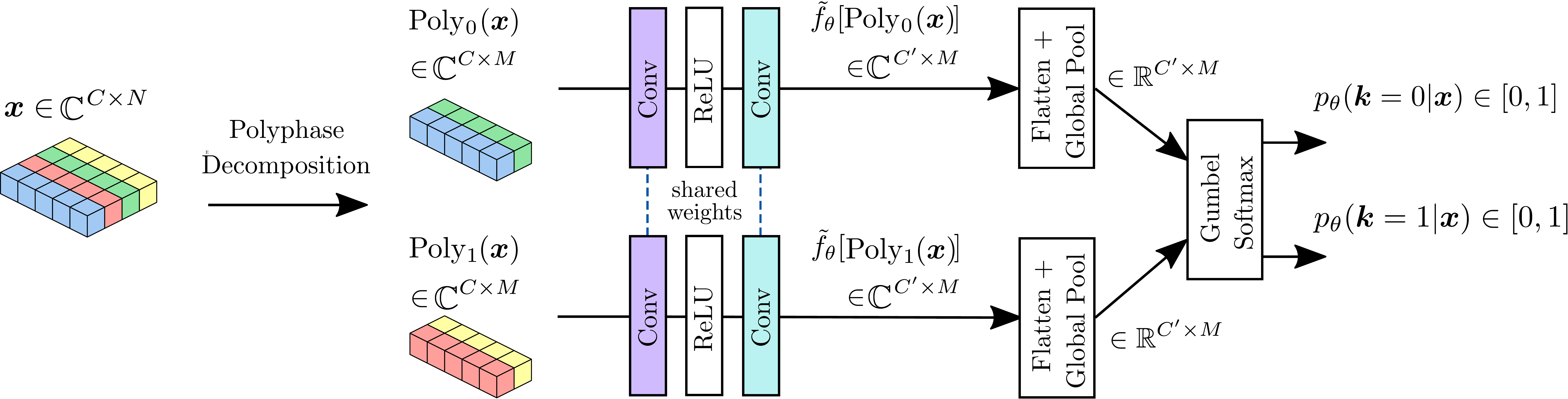}
    \caption{Proposed complex-valued extension to the shift-equivariant model first introduced in \cite{rojas2022learnable}. During training, we retain the weights sharing, pooling operations, and stochasticity as in \cite{rojas2022learnable}. We must project tensors to $\mathbb{R}$ before computing the Gumbel Softmax (or apply the Gumbel Softmax independently on the real and imaginary parts) to handle complex-valued weights and inputs.}
    \label{fig:lpd_complex}
\end{figure*}

This section presents a general methodology to define shift-invariant/equivariant methods to "make convolutional networks shift-equivariant" (regardless of their real or complex-valued nature). 
LPF aims to replace the traditional $\mathrm{MaxPool}$, strided convolution, and $\mathrm{AvgPool}$ operations with variants that utilize low-pass filters, thereby reducing the impact of these operations on shift-invariance. However, while LPF has tried to resolve this problem, it is not a shift-invariant method but rather a good approximation of this property. To address this limit, shift-invariant APS and LPS methods have been introduced.
\\
We now present the baseline methodology of shift-invariant methods. Let us begin by defining the polyphase component operator $\mathrm{Poly}$ acting on $\mathbb{C}^N$ as:
\begin{equation}
 \mathrm{Poly}_k\left(\mathbf{z}\right)[n] = \mathbf{z}[2n+k], \,\,  \forall n\in \left\llbracket 1, \left\lfloor N/2 \right\rfloor \right\rrbracket, \nonumber 
\end{equation}
where $k\in\{0,1\}$ and $\mathrm{Poly}_k(\mathbf{z}) \in \mathbb{C}^{\left\lfloor N/2 \right\rfloor}$ denotes the $k$-th \textit{polyphase component}. We also remind 
\begin{equation}
\label{eq:polyphase_component}
\mathrm{Poly}_k\left(\mathbf{T}_N \, \mathbf{z}\right)= \begin{cases}\mathrm{Poly}_1(\mathbf{z}) & \text { if } k=0\, , \\ \mathbf{T}_M\mathrm{Poly}_0(\mathbf{z}) & \text { if } k=1\, ,\end{cases}
\end{equation}
with $M=\left\lfloor N/2 \right\rfloor $.
Moreover, let us denote $p_{\mathbf{z}}(K|\mathbf{z})$ as the conditional probability for selecting \textit{polyphase components}, with $K$ denoting the Bernoulli random variable (a one-dimensional case of the categorical distribution) of polyphase indices and $\mathrm{PD}$ as a complex-valued Polyphase Downsampling layer, $\forall \mathbf{z} \in \mathbb{C}^N$,  
\begin{equation}
 \mathrm{PD}(\mathbf{z}) =\mathrm{Poly}_{k^{*}}\left(\mathbf{z}\right) \, , \nonumber 
\end{equation}
where $k^{*}=\displaystyle \arg\max_{k\in\{0,1\}} p_{\mathbf{z}}\left(K=k|\mathbf{z}\right)$. \\
\\
We can now propose \textbf{Claim 1}, \textbf{Claim 2}, and \textbf{Claim 3} which are extensions of existing real-valued claims \cite{rojas2022learnable}.\\
\\
\textbf{Claim 1:} \textit{If $p_{\mathbf{z}}$ is shift-permutation-equivariant, then $\mathrm{PD}$ is a shift-equivariant complex-valued downsampling layer.}\\
\\
We say that $p_{\mathbf{z}}$ is shift-permutation-equivariant iif:
\begin{equation}
p_{\mathbf{z}}\left(K=\pi(k)|\mathbf{T}_{N}\mathbf{z}\right)= p_{\mathbf{z}}(K=k|\mathbf{z})\, ,
\label{eq:claim1}
\end{equation}
where $\pi(.)$ denotes the permutation on the polyphase indices $\pi(k) = k +1 \mod p$ with $p$ the number of polyphase components (downsampling factor). When there is only 2 polyphase components, $p=2$, then  $\pi(1) = 0$ and  $\pi(0) = 1$.   In other words, shifting $\mathbf{z}$ leads to selecting the other polyphase component. The proof of this claim is provided in Appendix~\ref{app:proofs}.
\\
For instance, consider a typical downsampling scheme, such as the strided convolution. A convolution with stride $s$ is equivalent to a convolution with stride $1$ followed by a downsampling-by-$s$ operator (see Eq.~\eqref{eq:downsampling}). The downsampling operator~\eqref{eq:downsampling} does not follow equality~\eqref{eq:claim1} as soon as $s \geq 2$, since it is guaranteed that $K = s$, regardless of whether the data $\mathbf{z}$ may or may not have been circular-shifted. It follows that the usual strided convolution is not a shift-equivariant downsampling layer.
\\
One way to ensure that $p_\mathbf{z}$ is shift-equivariant is to define a shift-invariant $f:\mathbb{C}^{\left\lfloor N/2 \right\rfloor} \mapsto \mathbb{R}$ function, and an adaptive sampling layer $p_\mathbf{z}$ as the conditional probability:
\begin{align}
    \label{eq:claim2}
    p_\mathbf{z}(K=k|\mathbf{z})
    &=\frac{\exp{\left(f(\mathrm{Poly}_{k}(\mathbf{z}))\right)}}{\displaystyle\sum_{j \in \{0, 1\}} \exp{\left(f(\mathrm{Poly}_{j}(\mathbf{z}))\right)}}\, .
\end{align}
\textbf{Claim 2:} \textit{If $f$ is shift-invariant, then $p_\mathbf{z}$ is shift-permutation equivariant.}\\
\\
The proof of this claim is provided in Appendix~\ref{app:proofs}. Once $f$ is defined as a shift-invariant function, the actual choice of the function remains open. Many options have been proposed in the literature: the choice of function dramatically influences the neural network's performance.
For instance, APS proposed to define $f(\cdot)=||\cdot||$, implying that 
$\displaystyle k^*= \arg\max_{k\in\{0,1\}} ||\mathrm{Poly}_{k}(\mathbf{z})||$ since the $\mathrm{Softmax}$ operator is a smooth approximation of the $\arg \max$ operator on its input components. The polyphase component selection is based on the maximal norm value. Hence, as stated in Eq. \eqref{eq:invariance}, $f$ is shift-invariant (due to the global respective field of the max operator). As such, \textbf{Claim 1} and \textbf{Claim 2} are respected, proving that the downsampling layer proposed in APS is indeed shift-equivariant.
\\
While APS hardcoded the selection rule of the polyphase components, LPS introduced a learnable rule $f_{\boldsymbol{\theta}}:\mathbb{R}^{C\times H\times W}\mapsto\mathbb{R}$ as a small convolutional neural network parametrized by $\boldsymbol{\theta}$ that learn a downsampling scheme alongside the model training. This network does not use any downsampling scheme: $f_{\boldsymbol{\theta}}$ is composed of convolutional layers with a stride $1$ and a global pooling at the output, as seen in Figure \ref{fig:lpd_complex}. Hence, as stated in Eq. \eqref{eq:invariance}, $f_{\boldsymbol{\theta}}$ is shift-invariant. As such, \textbf{Claim 1} and \textbf{Claim 2} are respected, proving that the downsampling layer proposed in LPS is indeed shift-equivariant. We note that LPS generalizes the APS method (as demonstrated \cite[Claim 2]{rojas2022learnable}).
\\
Finally, to ensure that the shift-equivariance property is correctly validated during the upsampling (for segmentation and reconstruction tasks), we denote the partial inverse polyphase component (IPoly) operator acting on $\mathbf{C}^N$ as $\forall n\in \left\llbracket 1, N \right \rrbracket, \forall j \in \{0,1\}$:
\begin{equation}
     \mathrm{IPoly}_k(\mathbf{z})[2n+j] = \begin{cases}
        \mathbf{z}[n], \, \mathrm{if} \, j=k\\
        0, \, \mathrm{otherwise\, ,}
    \end{cases} 
\end{equation}
where $k\in\{0,1\}$ and $\mathrm{IPoly}(\mathbf{z})_k \in \mathbb{C}^{2N}$ denotes the $k$-th \textit{upsampled component}. Note that the IPoly operator satisfies the property:
$\mathrm{Poly}_k(\mathrm{IPoly}_k(\mathbf{z})) = \mathbf{z}$.
Given a downsampled feature map $\mathbf{y} = \mathrm{PD}(\mathbf{z}) \triangleq \mathrm{Poly}_{k^*}(\mathbf{z})$, the complex-valued polyphase upsampling layer is defined as $\mathrm{PU}(\mathbf{y}) =  \mathrm{IPoly}_{k^*}(\mathbf{y})$ where $k^{*}=\displaystyle \arg\max_{k\in\{0,1\}} p_{\mathbf{z}}\left(K=k|\mathbf{z}\right)$.
Thus, we propose the following claim:\\
\\
\textbf{Claim 3:} \textit{If $p_{\mathbf{z}}$ is shift-permutation equivariant, then $\mathrm{PU} \circ \mathrm{PD}$ is shift-equivariant.}\\
\\
A mandatory condition for \textbf{Claim 3} to be valid is that the overall architecture of the model is symmetric, i.e., the number of subsampling layers in the encoder must be equal to the number of upsampling layers in the decoder. The proof of this assertion is the extension of the real-valued proof provided in \cite[Claim 4]{rojas2024making}. The proof of Claim 3 is provided in Appendix~\ref{app:proofs}.
\\
While $\mathrm{PU}$ provides a shift-equivariant upsampling scheme, it introduces zeros in the output, which results in high-frequency components. This is known as aliasing in a multirate system \cite{vetterli2014foundations}. Similar to \cite{rojas2022learnable}, we apply a low-pass filter scaled by the upsampling factor after each PU to resolve this.
\\
In the following section, we propose to design a function $f_{\boldsymbol{\theta}}:\mathbb{C}^{C\times H\times W}\mapsto\mathbb{R}$ that generalizes the function $f_{\theta}$ proposed in APS and LPS.
\subsection{Learnable Projection Function} \label{sec:projection}
To effectively learn a downsampling scheme, we parametrize $f_{\theta}$ as a small complex-valued neural network using Gumbel Softmax \cite{maddison2016concrete, Gumbel17}. The Gumbel Softmax is a differentiable method that samples from a categorical distribution using a temperature parameter, allowing it to asymptotically converge to the $\arg \max$ operation. Mathematically, we define $\left[x_{i},\cdots,x_{n} \right]^T \in \mathbb{R}^{n}$  as the logits for selecting the polyphase components, $\left[g_{i},\cdots,g_{n}\right]$ as independent samples from standard Gumbel distribution. After integration, we obtain:
\begin{equation}
    \mathbb{P}\left(j=\arg \max_{i}(g_{i}+x_{i}) \right)=\frac{\exp{(x_{j})}}{\displaystyle \sum_{i=1}^n \exp{(x_{i})}}\, .
\label{eq:gumbel}
\end{equation}
A detailed introduction of the Gumbel Softmax method is provided in Appendix~\ref{app:gumbel}. Its extension to the complex domain requires specific attention. Logits being real-valued is a mandatory requirement for using the Gumbel Softmax, as it transforms a real-valued logit and a random Gumbel noise into a categorical distribution. To tackle this issue, we present a straightforward adaptive method to map logits from $\mathbb{C}^{N}$ to $\mathbb{R}^{N}$ in Section \ref{sec:projection}. \\
Due to the partial order of $\mathbb{C}$, extending existing operators is not straightforward. For instance, the Gumbel Softmax method requires a real-valued logit to be passed as an argument. Note that the Gumbel Softmax is not the only component that needs projecting complex-valued logits to $\mathbb{R}$; this is also the case for the cross-entropy loss function, as discussed in Section \ref{sec:preli}.  \\
As projecting from $\mathbb{C}^{N}$ to $\mathbb{R}^{N}$ can be done in various ways, we propose to distinguish between two types of projection: \textbf{implicit} and \textbf{explicit projection}.
\subsubsection{Implicit Projection}
For any $\mathbf{z} \in \mathbb{C}^N$, implicit projection is realized by applying a function on $\Im(\mathbf{z})$ and $\Re(\mathbf{z})$ separately and composing the results. When dealing with implicit projection, $\mathrm{Softmax}$ is a natural choice by computing the \textbf{Mean Softmax $\in [0,1]^{N}$} and  the \textbf{Product Softmax $\in [0,1]^{N}$} of $\mathrm{Softmax}(\Im(\mathbf{z}))$ and $\mathrm{Softmax}(\Re(\mathbf{z}))$ as: 
\begin{align*}
    \mathrm{MSoftmax}(\mathbf{z}) & =  \frac{1}{2}\left[\mathrm{Softmax}(\Re(\mathbf{z}))  + \mathrm{Softmax}(\Im(\mathbf{z}))\right] \, ,\nonumber \\
    \mathrm{PSoftmax}(\mathbf{z}) & =  \mathrm{Softmax}(\Re(\mathbf{z}))  \odot  \, \mathrm{Softmax}(\Im(\mathbf{z}))) \, . \nonumber
\end{align*}  
We naturally extend these two functions to obtain the equivalent for the Gumbel Softmax, i.e., $\mathrm{MGumbelSoftmax}$ and $\mathrm{PGumbelSoftmax}$.

\subsubsection{Explicit Projection}
Explicit projection consists of applying an additional function defined as $f:\mathbb{C}^{N}\mapsto\mathbb{R}^{N}$ on top of a neural network layer. The advantage of this approach over implicit projection is that it allows for more flexibility, due to its trainable nature. Computing the mean/product of probabilities, as done in the implicit projection approach, lacks a rigorous mathematical foundation, whereas defining a function that maps $\mathbb{C}^{N}$ to $\mathbb{R}^{N}$ is more principled.  
\\
\textbf{Norm:} For any $\mathbf{z} \in \mathbb{C}^N$, a well-known explicit projection function is the norm function, defined as $|\mathbf{z}|$. \footnote{Using the Norm function as a projection layer for the LPS method differs from the APS method. Indeed, APS selects components using the argmax on the norm of its input.} Nevertheless, while this function is commonly used to project a complex number into the real domain, the phase information is discarded. This payoff might be undesirable in many cases. To address this issue, we propose a learnable projection function that generalizes the norm function while leveraging the adaptability of the connectionist paradigm.
\\
\textbf{Polynomial:} We introduce a learnable projection polynomial layer to generalize the norm function by adapting the projection function during the learning process. The proposed layer processes complex-valued input data $\mathbf{z} \in \mathbb{C}^N$ by decomposing each component $\mathbf{z}_l$ of the vector $\mathbf{z}\in \mathbb{C}^N$, for $l\in \llbracket 1,N \rrbracket$, into a $M$-polynomial function acting on its real and imaginary components, $\mathbf{a}_l = \Re(\mathbf{z})_l$ and $\mathbf{b}_l = \Im(\mathbf{z})_l$, respectively. It computes all terms $\mathbf{a}_l^{i} \, \mathbf{b}_l^{j}$ where $i, j \geq 0 $, $ i + j = m$, for each degree $m$ ranging from 1 to $M$, the specified order of the polynomial expansion. These terms are used as inputs to a linear transformation, resulting in the output, $\forall l\in \left\llbracket 1, N \right\rrbracket$:
\begin{equation}
    \mathrm{PolyDec}_l(\mathbf{z}) = \beta +  \sum_{m=1}^{M} \sum_{i=0}^{m} \theta_{m,i} \, \mathbf{a}_l^{i} \, \mathbf{b}_l^{m - i},
\end{equation}
where $\left\{\theta_{i, j}  \in \mathbb{R} \right\}_{(i, j)\in [1, M]^2}$ are the learned weights and $\beta \in \mathbb{R}$ is the bias term of the linear layer which is for the constant term ($i = j = 0$). This formulation effectively models $f_{\boldsymbol{\theta}}$ as a polynomial function $\mathrm{PolyDec}$ of the real and imaginary parts of $\mathbf{z}$, up to degree $M$, capturing complex interactions between them.
\\
\\
\textbf{MLP:} While the amplitude is a hard-wired function and the polynomial projection function is constrained to follow an explicit formulation, we also introduce a more general class of learnable projection functions as an MLP. We define the multilayer perceptron (MLP) as a sequence of fully connected layers that process the real and imaginary parts of the complex input $\mathbf{z}$, such as:
\begin{equation}
    \mathrm{MLP}(\mathbf{z}) = f^{(L)} \circ f^{(L-1)} \circ \ldots\circ f^{(1)} \left( \begin{bmatrix} \Re(\mathbf{z}) \\[1pt] \Im(\mathbf{z}) \end{bmatrix} \right),
\end{equation}
where $L$ is the number of layers in the network, each layer $f^{(l)}$ for $l \in \left\llbracket 1, L \right\rrbracket$ is defined as a real-valued non-linear transformation.
\\
Based on the universal approximation theorem, this layer can generalize all possible projection functions without being constrained to a specific polynomial form. Nevertheless, note that its size constrains the generalization capabilities of the $\mathrm{MLP}$.
\section{Experiments}
\label{sec:experiments}
In this section, we evaluate the performance of complex-valued shift-invariant/equivariant neural networks in the complex domain. More precisely, we demonstrate how complex-valued shift-invariant/equivariant neural networks outperform traditional complex-valued neural networks across various computer vision tasks and possess an inherent superiority in their design, particularly in terms of the impact of the shift on the input. We also test our approach on dual-RVNNs to evaluate the effect of the complex nature of CVNNs compared to their real-valued counterparts.
We conduct experiments on image classification, semantic segmentation, and image reconstruction. To align our experiments with the scope of this article, we carefully selected complex-valued imaging datasets upon which our models were tested. Indeed, by only experimenting to the data's physical background. While we reproduce the existing setup for circular shift \cite{chaman2021truly}, we propose the first setup to evaluate complex-valued shift-invariant/equivariant neural networks in this section. More precisely, we used the Cr. S. metric from \cite{chaman2021truly, zhang2019making, rojas2022learnable} to evaluate shift-equivariance/invariance by applying a random circular shift between 1 and 9 on the input of a model, and its output. Shift-equivariant/invariant models should achieve a $100\%$ perfect score, unlike non-equivariant/invariant models. 
\\
However, we note that outperforming existing state-of-the-art models is outside the scope of this work, as we primarily focus on assessing the exact impact of each component of our method. To this end, we tested our methods on complex-valued adaptations of standard SOTA CNNs architectures (ResNet \cite{he2016deep}, UNet \cite{ronneberger2015u}, AutoEncoder\cite{hinton1993autoencoders}).
We detail each task's architectures and training setup in Appendix \ref{app:setup} and provide a complete guide to reproduce our experiments at the following repository\footnote{\url{https://github.com/QuentinGABOT/Equivariant_CVNN}}.
\subsection{Datasets}
While experiments presented in this section are based on computer vision tasks, the dataset of interest concerns polarimetric SAR (PolSAR)  imagery \cite{Pottier09}. This imaging technique is based on the coherent combination of radar echoes, yielding an image of the ground scene. The SAR system is usually mounted on a satellite or an aircraft, providing complex-valued information about the backscatter characteristics of the ground. Each pixel of a PolSAR image is associated with a complex scattering matrix $\mathbf{S}$, called the Sinclair matrix $\mathbf{S} = \begin{pmatrix}
 S_{HH} & S_{VH} \\
 S_{HV} & S_{VV} \\
 \end{pmatrix}$ where the indices $H$ and $V$ refer to the horizontal and vertical polarization states, but any orthogonal polarimetry basis can be used. PolSAR datasets are presented in more detail in Appendix \ref{app:datasets}, while polarimetric decompositions are detailed in Appendix \ref{app:polar}.
\\
Note that even if the following results have been obtained with PolSAR images, other types of complex-valued imaging could have been used, notably magnetic resonance imaging (MRI) \cite{liang2000principles}. We specifically recognize that CVNNs have been utilized for specific MRI applications, such as fat-water separation and flow quantification \cite{cole2021analysis}.
\begin{table}[htbp]
\centering
\resizebox{\linewidth}{!}{
\begin{tabular}{|c|c|c|c|c|c|c|c|c|c|c|c|}
\hline
Model & Type & OA (\%) & F1 (\%) & Cr. S. (\%) \\
\hline
ResNet LPS $\mathrm{MLP}$ & $\mathbb{C}$ & $\mathbf{84.97}$ & $\mathbf{80.62}$ & $\mathbf{100.0}$ \\
\hline
ResNet LPS & $\mathbb{R}$ & $82.92$ & $74.49$ & $\mathbf{100.0}$ \\
\hline
ResNet LPS $\mathrm{PolyDec}$  & $\mathbb{C}$ & $81.63$ & $73.5$ & $\mathbf{100.0}$ \\
\hline
ResNet LPS $\mathrm{MSoftmax}$ & $\mathbb{C}$ & $79.5$ & $72.23$ & $\mathbf{100.0}$ \\
\hline
ResNet LPS $\mathrm{Norm}$ & $\mathbb{C}$ & $76.38$ & $68.25$ & $\mathbf{100.0}$ \\
\hline
ResNet APS & $\mathbb{C}$ & $82.01$ & $75.37$ & $\mathbf{100.0}$ \\
\hline
ResNet APS & $\mathbb{R}$ & $81.16$ & $73.08$ & $\mathbf{100.0}$ \\
\hline
\hline\hline
ResNet LPF & $\mathbb{C}$ & $81.56$ & $73.68$ & $93.5$ \\
\hline
ResNet LPF & $\mathbb{R}$ & $79.66$ & $70.8$ & $95.44$ \\
\hline
\end{tabular}
}
\caption{Classification and circular shifts metrics on the \text{S1SLC\_CVDL} dataset. Shift-equivariant models are separated from non-shift-equivariant models, namely CVNN/RVNN LowPass Filter (LPF), by the additional row. CVNN Learnable Polyphase Sampling (LPS) with $\mathrm{MLP}$ projection layer outperforms all variants, even Adaptive Polyphase Sampling (APS).}
\label{tab:s1slc}
\end{table}
\subsection{Classification}
\label{exp:classification}
To test our approach on the classification task, we propose an experiment setup based on the \text{S1SLC\_CVDL} dataset, as described in Appendix \ref{app:setup_classif}. On average, we report for an A100 GPU, the following training times per epoch: CVNN LPS $\mathrm{PolyDec}$: $32m30s$, CVNN LPF: $22m15s$, RVNN LPF: $06m06s$. We also report the number of trainable parameters and the inference time for the shift-equivariant CVNN variants in Appendix~\ref{app:setup_classif}. Results from Table \ref{tab:s1slc} validate our initial hypothesis that complex-valued shift-invariant CVNNs outperform their non-shift-invariant counterparts.
\\
Our proposed approach, a CVNN LPS with an $\mathrm{MLP}$ projection function, outperforms every other model. Unlike non-shift-invariant models, shift-invariant CVNNs and RVNNs score a perfect circular shift consistency. Furthermore, the three best-performing models (when looking at the Overall Accuracy (OA) and the F1 score) are shift-invariant, highlighting this property's benefits.
Nevertheless, we note that not \textit{all} shift-invariant models outperform non-shift-invariant models based on the OA and F1 score: this interesting result likely indicates that the design choice behind the downsampling scheme and the projection function matters. More specifically, the APS approach consistently appears to be a worse choice than the LPS strategy, indicating that this non-learnable downsampling scheme is outperformed by its learnable variant. This observation can also be made when comparing our different projection functions: the $\mathrm{Norm}$ function is also generalized and outperformed by the $\mathrm{PolyDec}$ and the $\mathrm{MLP}$.
Finally, we observe that RVNNs \textit{almost always} underperformed against their CVNN counterparts. We believe this result comes from the arduous classification of classes from the \text{S1SLC\_CVDL} dataset. Naturally, exploiting the phase information might confer CVNNs a unique advantage over RVNNs. 
We provide additional results and visualizations in Appendix \ref{app:more_results_classif}.
\begin{table}[htbp]
\centering
\resizebox{\linewidth}{!}{
\begin{tabular}{|c|c|c|c|c|c|c|c|c|c|c|c|}
\hline
Model & Type & OA (\%) & F1 (\%) & Cr. S. (\%) \\
\hline
UNet LPS $\mathrm{PolyDec}$ & $\mathbb{C}$ & $\mathbf{97.21}$ & $\mathbf{92.7}$ & $\mathbf{100.0}$ \\
\hline
UNet LPS & $\mathbb{R}$ & $\mathbf{97.21}$ & $91.86$ & $\mathbf{100.0}$ \\
\hline
UNet LPS $\mathrm{Norm}$ & $\mathbb{C}$ & $96.04$ & $89.47$ & $\mathbf{100.0}$ \\
\hline
UNet LPS $\mathrm{MLP}$  & $\mathbb{C}$ & $95.24$ & $89.32$ & $\mathbf{100.0}$ \\
\hline
UNet LPS $\mathrm{MSoftmax}$ & $\mathbb{C}$ & $94.59$ & $87.05$ & $\mathbf{100.0}$ \\
\hline
UNet APS  & $\mathbb{R}$ & $96.73$ & $90.15$ & $\mathbf{100.0}$ \\
\hline
UNet APS & $\mathbb{C}$ & $93.78$ & $83.85$ & $\mathbf{100.0}$ \\
\hline
\hline\hline
UNet LPF & $\mathbb{R}$ & $97.13$ & $91.62$ & $97.01$ \\
\hline
UNet LPF & $\mathbb{C}$ & $93.12$ & $84.58$ & $97.04$ \\
\hline
\end{tabular}
}
\caption{Semantic segmentation and circular shifts metrics on the PolSF dataset. Shift-equivariant models are separated from non-shift-equivariant models, namely CVNN/RVNN Low-Pass Filter (LPF), by the additional row. CVNN Learnable Polyphase Sampling (LPS) with $\mathrm{PolyDec}$ projection layer outperforms all variants, even Adaptive Polyphase Sampling (APS).}
\label{tab:polsf}
\end{table}
\subsection{Semantic Segmentation}
\label{exp:segmentation}
We propose an experiment setup based on the PolSF dataset, as described in Appendix \ref{app:setup_seg}, to test our approach to the semantic segmentation task. On average, we report the following training times per epoch for the A100 GPU: CVNN LPS $\mathrm{PolyDec}$: $ 38$ seconds, CVNN LPF: $ 34$ seconds, RVNN LPF: $ 14$ seconds. We also report the number of trainable parameters and the inference time for the shift-equivariant CVNN variants in Appendix~\ref{app:setup_seg}. Table~\ref{tab:polsf} shows that the results validate our initial hypothesis that complex-valued shift-equivariant CVNNs outperform their non-shift-equivariant counterparts.
\\
Results show that the proposed approach outperforms all other models, specifically a CVNN LPS with a $\mathrm{PolyDec}$ projection function. Unlike non-shift-equivariant models, shift-equivariant CVNNs and RVNNs score a perfect circular shift consistency. Furthermore, the two best-performing models are shift-equivariant with respect to performance metrics, like OA and F1 score.
Nevertheless, we note that not \textit{all} shift-equivariant models outperform non-shift-equivariant models based on the OA and F1 score; thus, we reiterate the same conclusions as in Section \ref{exp:classification}.
Finally, we observe that RVNNs \textit{almost always} outperform their CVNN counterparts. Contrary to the observations made in Section \ref{exp:classification}, we believe that this might be because the semantic segmentation of the PolSF dataset is a rather simple task (notably due to the relatively small size of the dataset): the phase information might not be a sufficient argument to use CVNNs over RVNNs in this case. Nonetheless, the CVNN LPS $\mathrm{PolyDec}$ appears to be a striking exception to this observation: this leads us to hypothesize that CVNNs may have a more challenging time learning some representations than RVNNs, but they ultimately possess a more substantial representation capability if correctly designed. 
We provide additional results and visualizations in Appendix \ref{app:more_results_seg}.
\subsection{Reconstruction}
\label{exp:reconstruction}
Finally, to test our approach on the reconstruction task, we propose an experiment setup based on the San Francisco Polarimetric SAR ALOS-2 dataset, as described in Appendix \ref{app:setup_recon}. The polarimetric properties of the data allow us to evaluate the reconstruction using additional metrics by computing coherent and non-coherent polarimetric decompositions on the reconstructed image. We leverage these decompositions to assess the model's performances: namely, the Pauli \cite{cloude1996review, lee2017polarimetric}, Krogager \cite{Krogager92, Krogager95, Krogager95A}, Cameron \cite{cameron1990feature, cameron2002simulated, cameron2006conservative}, and $H-\alpha$ \cite{cloude2002entropy, formont2010statistical} decompositions, as presented in Appendix \ref{app:polar}. Then, as is usually done by the signal processing community, we compute classification metrics from the Cameron decomposition and the $H-\alpha$ decomposition. Please note that the following metrics are obtained from an unsupervised process meant to be calculated on a fully trained model on the reconstruction task \cite{gabotpreserving}. More details regarding the experiments can be found in Appendix \ref{app:more_results_recon}. On average, we report the following training times per epoch for the A100 GPU: CVNN LPS $\mathrm{PolyDec}$: $ 22$ seconds, CVNN LPF: $ 13$ seconds, RVNN LPF: $ 2$ seconds. We also report the number of trainable parameters and the inference time for the shift-equivariant CVNN variants in Appendix~\ref{app:setup_recon}. Finally, we did not apply the low-pass filter for the reconstruction task, as it does not show any improvements.
\begin{table}[htbp]
\centering
\resizebox{\linewidth}{!}{
\begin{tabular}{|c|c|c|c|c|c|c|c|c|c|c|c|}
\hline
Model & Type & MSE $(\downarrow)$ & H-$\alpha$ OA (\%) & H-$\alpha$ F1 (\%) & Cam. OA (\%) & Cam. F1 (\%) & Cr. S $(\downarrow)$\\
\hline
AE LPS $\mathrm{PolyDec}$ & $\mathbb{C}$ & $\mathbf{1.3\times10^{-4}}$ & $\mathbf{96.66}$ & $\mathbf{95.56}$ & $\mathbf{94.03}$ & $\mathbf{91.40}$ & $\mathbf{0.0}$\\
\hline
AE APS & $\mathbb{C}$ & $1.5\times10^{-4}$ & $95.50$ & $93.88$ & $93.05$ & $90.23$ & $\mathbf{0.0}$ \\
\hline
AE LPS $\mathrm{MLP}$ & $\mathbb{C}$ & $3.6\times10^{-4}$ & $94.54$ & $92.67$ & $90.41$ & $86.74$ & $\mathbf{0.0}$ \\
\hline
AE LPS $\mathrm{Norm}$ & $\mathbb{C}$ & $4.2\times10^{-4}$ & $94.33$ & $92.37$ & $89.62$ & $86.04$ & $\mathbf{0.0}$ \\
\hline
AE APS & $\mathbb{R}$ & $1.1\times10^{-2}$ & $60.53$ & $51.13$ & $58.40$ & $45.56$ & $\mathbf{0.0}$ \\
\hline
AE LPS $\mathrm{MSoftmax}$ & $\mathbb{C}$ & $1.2\times10^{-2}$ & $58.05$ & $49.17$ & $59.17$ & $50.87$ & $\mathbf{0.0}$ \\
\hline
AE LPS & $\mathbb{R}$ & $2.2\times10^{-2}$ & $53.38$ & $42.20$ & $46.03$ & $34.40$ & $\mathbf{0.0}$ \\
\hline
\hline\hline
AE LPF & $\mathbb{C}$ & $6.3\times10^{-2}$ & $46.97$ & $32.69$ & $23.74$ & $17.01$ & $72.21$ \\
\hline
AE LPF & $\mathbb{R}$ & $8.5\times10^{-2}$ & $34.15$ & $19.01$ & $16.26$ & $11.33$ & $43.12$ \\
\hline
\end{tabular}
}
\caption{Reconstruction, $H-\alpha$ classification metrics, and Cameron classification metrics on the San Francisco Polarimetric SAR ALOS-2 dataset. Shift-equivariant models are separated from non-shift-equivariant models, namely CVNN/RVNN Low-Pass Filter (LPF), by the additional row. CVNN Learnable Polyphase Sampling (LPS) with $\mathrm{PolyDec}$ projection layer outperforms all variants, even Adaptive Polyphase Sampling (APS). Note that the Cr. S metric differs from the supervised learning task, as we cannot measure the agreement between the predictions. As such, we compute the $\ell_2$ norm: in this case, shift-equivariant models achieve a perfect score, while non-shift-equivariant models obtain non-zero values.}
\label{tab:alos2}
\end{table}
Results from Table \ref{tab:alos2} validate our initial hypothesis that complex-valued shift-equivariant CVNNs outperform their non-shift-equivariant counterparts. We can also notice the pattern observed in Section \ref{exp:classification}, i.e., that CVNNs outperform their RVNN counterparts. Even though we believe these results are logical, as polarimetric decompositions (such as $H-\alpha$) are highly sensitive to amplitude and phase values, these results highlight that the CVNNs are particularly well-suited for the reconstruction task, which is even more accentuated when they are shift-equivariant. Furthermore, we notice a clear distinction in performance between the shift-equivariant models (the first four, as opposed to the last three), further validating our hypothesis that the choice of both downsampling and projection layers matters.
\section{Conclusion} 
\label{sec:discussions}
This work extended a provably shift-equivariant convolutional neural network framework to the complex domain. In doing so, we enabled the incorporation of this strong prior to convolutional CNNs.
In our experiments, we observed that LPS demonstrated better results than APS, demonstrating that a learnable downsampling scheme has an advantage over a handcrafted one.
Following this observation, we introduced learnable functions  $\mathrm{PolyDec}$ and $\mathrm{MLP}$ to perform the needed projection from $\mathbb{C}$ to $\mathbb{R}$.
This end-to-end trainable approach allows the model to adapt effectively to the data it is trained on. 
In our experiments, we observe that the $\mathrm{PolyDec}$ projection function, and even more so, the $\mathrm{MLP}$ projection function, increase computational requirements and latency.
This is even more aggravating when dealing with CVNNs, as they are slower to train than their RVNN counterparts.
Thus, we believe the $\mathrm{PolyDec}$ projection function strikes a sweet intermediate between a non-learnable projection function, like $\mathrm{Norm}$, and an unrestricted one, like $\mathrm{MLP}$.
\clearpage
{
    \small
    \bibliographystyle{ieeenat_fullname}
    \bibliography{main}
}
\clearpage
\appendix

\appendix

\section*{Appendix}

\section{Impact of Stride on Shift-Equivariance}
\label{app:shift}
To understand why the stride is the root of the issue regarding shift-equivariance, let the downsampling be reformulated as an operator (e.g., max, average, convolution) applied with a stride of $1$ followed by subsampling depending on the subsampling factor. In this case, the subsampling scheme is hard-wired and returns the values at the indices modulo the scale factor. For instance, consider a 1D signal $f[n] = n$ and a max-pooling operation with a window size of $2$. We first compute the stride-$1$ pooled signal: $g[n]=\max\bigl(f[n],\,f[n+1]\bigr)$, and then sub-sample by taking every other value:
\begin{equation*}
      s[k] = g[2k]
       = \bigl[\max(f[0],f[1]),\,,\dots\bigr]
       = [1,3,\dots]
\end{equation*}
By shifting the input by one, we get $f'[n] = f[n+1]$, which yields
\begin{equation*}
    s'[k] = g'[2k]
        = \bigl[\max(f'[0],f'[1]),\,\dots\bigr]
        = [2,4,\dots]
\end{equation*}
If pooling were shift‐equivariant we would have $s'[k]=s[k+1]$, but $s'[0]=2\neq3=s[1]$, demonstrating that fixed-index sub-sampling breaks shift‐equivariance. This fixed strategy, independent of the input representation, cannot be shift-invariant/equivariant.
\section{Proofs of Claim 1-3}
\label{app:proofs}
\paragraph{Claim 1} The following proof is an extension of the real-valued demonstration provided in \cite[Claim 1]{rojas2022learnable}.
\begin{proof}
    Let $\hat{\mathbf{z}} \triangleq \mathbf{T}_{N}\mathbf{z}$ be a shifted version of $\mathbf{z}\in\mathbb{C}^{N}$. Recall $\mathrm{PD}(\mathbf{z})$ and $\mathrm{PD}(\hat{\mathbf{z}})$ are defined as:
    \begin{align}
     \mathrm{PD}(\mathbf{z}) &\triangleq  \mathrm{Poly}_{k^{*}}(\mathbf{z}) \\
     \mathrm{PD}(\hat{\mathbf{z}}) &\triangleq  \mathrm{Poly}_{\hat{k}^{*}}(\hat{\mathbf{z}})
    \end{align}
    where $k^{*}=\displaystyle \arg\max_{k\in\{0,1\}} p_{\mathbf{z}}\left(K=k|\mathbf{z}\right)$, and $\hat{k}^{*}=\displaystyle \arg\max_{k\in\{0,1\}} p_{\mathbf{z}}\left(K=k|\hat{\mathbf{z}}\right)$. Claim 1 states that $p_{\mathbf{z}}$ is shift-permutation-equivariant i.e. that $\hat{k}^{*} = \pi(k^{*}) = 1 - k^{*}$. So, from Eq.~\eqref{eq:polyphase_component}, we show that:
    \begin{align}
     \mathrm{PD}\left(\mathbf{T}_N \, \mathbf{z}\right)&= \begin{cases}\mathrm{Poly}_1(\mathbf{z}) & \text { if } k^{*}=1\, , \nonumber \\ 
     \mathbf{T}_M\mathrm{Poly}_0(\mathbf{z}) & \text { if } k^{*}=0\, .\end{cases} \\ 
     &=((1-k^{*})\mathbf{T}_M+k^{*}\mathbf{I})\cdot\mathrm{PD}(\mathbf{z}) \label{eq:PDequivariant}
    \end{align}
    with $M=\lfloor N/2\rfloor$,  showing that PD satisfies the shift-equivariance definition.
\end{proof}
\paragraph{Claim 2} The following proof is an extension of the real-valued demonstration provided in \cite[Claim 3]{rojas2022learnable}.
\begin{proof}
    Denote a feature map $\mathbf{z}$ and its shifted version $\hat{\mathbf{z}} \triangleq \mathbf{T}_N\mathbf{z}$, such as using Eq.~\eqref{eq:claim2} we got:
    \begin{equation}
        p_\mathbf{z}(K=\pi(k)|\mathbf{T}_N\mathbf{z})
        =\frac{\exp{\left(f(\mathrm{Poly}_{k}(\mathbf{T}_N\mathbf{z}))\right)}}{\displaystyle\sum_{j \in \{0, 1\}} \exp{\left(f(\mathrm{Poly}_{j}(\mathbf{T}_N\mathbf{z}))\right)}}\, .
    \end{equation}
    From Eq.~\eqref{eq:polyphase_component} and given $f$ shift-invariant, we obtain:
    \begin{equation}
         f(\mathrm{Poly}_{\pi(k)}(\mathbf{T}_N\mathbf{z})) = f(\mathbf{T}_M\mathrm{Poly}_{k}(\mathbf{z}))=f(\mathrm{Poly}_k(\mathbf{z}))
    \end{equation}
    Finally, we have:
    \begin{align}
        p_\mathbf{z}(K=\pi(k)|\mathbf{T}_N\mathbf{z})
        &=\frac{\exp{\left(f(\mathrm{Poly}_{k}(\mathbf{z}))\right)}}{\displaystyle\sum_{j \in \{0, 1\}} \exp{\left(f(\mathrm{Poly}_{j}(\mathbf{z}))\right)}} \\
        &= p_\mathbf{z}(K=k|\mathbf{z})
    \end{align}
\end{proof}
\paragraph{Claim 3} The following proof is inspired by the real-valued demonstration provided in \cite[Claim 4]{rojas2022learnable}. Nevertheless, we provide a simpler and more general demonstration.
\begin{proof}
First, let $N=2M$, note that for any vector $\mathbf{y} \in \mathbb{C}^M$,
\begin{align}
    \mathrm{IPoly}_1(\mathbf{T}_M\mathbf{y})  &= \mathbf{T}_{N}\mathrm{IPoly}_0(\mathbf{y})  \\\mathrm{IPoly}_0(\mathbf{y}) &= \mathbf{T}_{N}\mathrm{IPoly}_1(\mathbf{y})
\end{align}
Then, denote a feature map $\mathbf{z}$ and its shifted version $\hat{\mathbf{z}} \triangleq \mathbf{T}_N\mathbf{z}$, and $\displaystyle k^* = \arg\max_{k\in\{0, 1\}}p_\theta(K=k|\mathbf{z})$. We define, $\mathbf{u} \triangleq  \mathrm{PU} \circ  \mathrm{PD}(\mathbf{z}) $ and   $\hat{\mathbf{u}} \triangleq  \mathrm{PU} \circ  \mathrm{PD}(\mathbf{T}_N\mathbf{z})$. Since $p_\mathbf{z}$ is shift-permutation-equivariant, the selection index for  $\mathbf{T}_N\mathbf{z}$ is switched to $\hat{k}^{*} = \pi(k^{*}) = 1 - k^{*}$.
From \cref{eq:PDequivariant}, we have,
\begin{equation}
    \hat{\mathbf{u}} = \mathrm{IPoly}_{\hat{k}^{*}}\left(((1-k^{*})\mathbf{T}_M+k^{*}\mathbf{I})\cdot\mathrm{PD}(\mathbf{z})\right)
\end{equation}
In the case $k^* = 0$, we have,
\begin{align}
    \hat{\mathbf{u}} &= \mathrm{IPoly}_{1}\left(\mathbf{T}_M\mathrm{Poly}_0(\mathbf{z})  \right)  = \mathbf{T}_N\mathrm{IPoly}_{0}\left(\mathrm{Poly}_0(\mathbf{z})  \right) \nonumber\\
    &=\mathbf{T}_N \mathbf{u},
\end{align}
and for the case $k^* = 1$,
\begin{align}
    \hat{\mathbf{u}} &= \mathrm{IPoly}_{0}\left(\mathrm{Poly}_1(\mathbf{z})  \right)
    = \mathbf{T}_N \mathrm{IPoly}_{1}\left(\mathrm{Poly}_1(\mathbf{z})  \right) \nonumber \\
    &= \mathbf{T}_N \mathbf{u},
\end{align}
showing that $\mathrm{PU} \circ  \mathrm{PD}$ is shift-equivariant.
\end{proof}
\section{Gumbel Softmax}
\label{app:gumbel}
We first provide a step-by-step integration process to highlight how the Gumbel Max-Trick works \cite{maddison2015asampling, gumbel1954statistical}.
We remind that the Cumulative Density Function (CDF) of the standard Gumbel distribution is
$F(z) =  \exp\left(-\exp(-z))\right)$ and its corresponding Probability Density Function (PDF) is $f(z) =  \exp\left(-z - \exp(-z)\right)$. \\
\\
The probability $\mathbb{P}\left(j = \arg\max_i (g_i + \log \pi_i)\right)$ defined in Eq. \eqref{eq:gumbel} where $x_i=\log \pi_i$ can be rewritten as:
\begin{equation}
    \prod_{i \neq j} \mathbb{P}\left(g_j + \log \pi_j > g_i + \log \pi_i\right)\, .
\end{equation}
This probability can be expressed with the Gumbel PDF $f(.)$, where $g_j = t$:  
\begin{equation}
    \int_{-\infty}^{\infty} \prod_{i \neq j} \mathbb{P}\left(g_i  < t + \log (\pi_j/\pi_i)\right) \, f(t) \, dt\, .
\end{equation}
Using the Gumbel CDF, $F(.)$, we obtain:
\begin{equation}
    \int_{-\infty}^{\infty} \prod_{i \neq j} \exp\left(-\exp\left(-t - \log (\pi_j / \pi_i)\right)\right) \, f(t) \, dt\, ,
\end{equation}
that can be rewritten as:
\begin{equation}
    \int_{-\infty}^{\infty} \exp\left(-\sum_{i \neq j} \frac{\pi_i}{\pi_j} \, \exp(-t)  \right) \, f(t) \, dt\, .
\end{equation}
Replacing the Gumbel PDF leads to the following:
\begin{equation}
     \int_{-\infty}^{\infty} \exp\left(- \frac{\exp(-t)}{\pi_j} \sum_{i \neq j} \pi_i\right) \, \exp\left(-t - \exp(-t)\right) \, dt\, .
\end{equation}
Rearranging terms gives:
\begin{equation}
    \int_{-\infty}^{\infty} \exp\left(-t - \displaystyle \frac{1}{\pi_j} \displaystyle\sum_{i} \pi_i \, \exp(-t) \right) \, dt\, .
\end{equation}
The following change of variable $y=-\displaystyle\frac{1}{\pi_j} \displaystyle\sum_{i} \pi_i \, \exp(-t)$ leads to the final results:
\begin{equation}
    \mathbb{P}\left(j = \arg\max_i (g_i + \log \pi_i)\right) = \displaystyle\frac{\pi_j}{\displaystyle\sum_{i} \pi_i} \, .
\end{equation}
The Gumbel-Max Trick allows for sampling from a categorical distribution during the forward pass through a neural network, as $z = \mathrm{one\_hot}\left(\displaystyle\arg\max_{i}[g_{i}+\log\alpha_{i}]\right)\sim \mathrm{Categorial}\left(\displaystyle\frac{\alpha_{i}}{\sum_{j}\alpha_{j}}\right)$ \cite{maddison2016concrete}. Nevertheless, since the $\arg\max$ function is not differentiable, we replace it with the $\mathrm{Softmax}$ function. A temperature factor $\lambda$ allows us to control how closely the Gumbel-softmax distribution approximates the categorical distribution \cite{maddison2016concrete}.
\section{Polarimetric Decomposition}
\label{app:polar}
\subsection{Coherent Polarimetric Decomposition}
A common tool for studying the Sinclair matrix $\mathbf{S}$ is based on coherent decompositions, which involve of canonical objects. The underlying reason is that significant dispersion and anisotropy phenomena are expected from coherent (or artificial) targets (e.g., cars, airplanes, trucks, buildings), unlike non-deterministic targets such as vegetation (e.g., crops, forests). Therefore, these decompositions can help us to distinguish them better. Two well-known representations are usually used: the Pauli and the Krogager decomposition \cite{Pottier09}.
\subsubsection{Pauli decomposition}
\label{sec:pauli}
In the context of monostatic SAR, the Pauli decomposition \cite{cloude1996review, lee2017polarimetric} expresses $\mathbf{S}$ as:
\begin{equation}
     \mathbf{S} =  \frac{\alpha}{2} \begin{pmatrix} 1 & 0 \\ 0 & 1 \end{pmatrix} + \frac{ \beta}{2} \begin{pmatrix} 1 & 0 \\ 0 & -1 \end{pmatrix} +  \frac{\gamma}{2} \begin{pmatrix} 0 & 1 \\ 1 & 0 \end{pmatrix},\,
     \label{eq:lexico}
\end{equation}
where each term $\alpha = S_{HH} + S_{VV}$, $\beta = S_{HH} - S_{VV}$ and $\gamma = 2S_{HV}$ represent the part of the response of a plate observed at normal incidence or a sphere, the characteristic of a horizontal metallic dihedral and the scattering matrix of a metallic dihedral oriented at $45^\circ$ with respect to the radar line of sight respectively. Then, we define the Pauli vector as $\displaystyle \mathbf{k} = \frac{1}{\sqrt{2}} \left( \alpha, \beta, \gamma \right)^T$ which is usually used to estimate the coherence matrix in Eq. (\ref{eq:coherence}) and to plot an RGB image by taking the module of each component.
\subsubsection{Krogager Decomposition}
A refined alternative approach, proposed by \cite{Krogager92,Krogager95, Krogager95A} considers a scattering matrix as the combination of the responses of a sphere, an oriented diplane, and a helix:
\begin{equation}
  \mathbf{S}=e^{j \varphi}\left(e^{j \varphi_s}\,k_s\,\mathbf{S}_{s}+k_d\,\mathbf{S}_{d}
    (\vartheta)+k_h\,\mathbf{S}_{h} (\vartheta)\right)\, ,
  \label{eq:krog_1}
\end{equation}
where $\mathbf{S}_{s}=\begin{pmatrix}  1 & 0\\0 &
      1\end{pmatrix}$,  $\mathbf{S}_{d} (\vartheta)=\begin{pmatrix}
      \cos 2\vartheta & \sin 2\vartheta\\\sin 2\vartheta &
      -\cos 2\vartheta \end{pmatrix}$ and $\mathbf{S}_{h} (\vartheta)=e^{\pm 2 j
    \vartheta}\begin{pmatrix}  1 & \pm j\\ \pm j &
      -1\end{pmatrix}$, 
where the $\pm$ sign in the helix component varies the left or
right-handedness, and it must be accounted for during the estimation of its components. The identification of the parameters is generally performed in the right-left circular basis:
\begin{equation}
    \begin{cases}
    k_s = |S_{RL}|\, , \\
    k_h =|S_{RR}|-|S_{LL}|, \, k_d =|S_{LL}| \quad \text{if } |S_{RR}| > |S_{LL}| \, ,\\
    k_h =|S_{LL}|-|S_{RR}|, \, k_d =|S_{RR}| \quad \text{otherwise.}
    \end{cases}
\end{equation}
The condition $|S_{RR}| > |S_{LL}|$ denotes the presence of a left-handed helix contribution. The coefficients $k_s$, $k_d$, and $k_h$ represent the amplitude of each canonical scattering mechanism contributing to the initially measured scattering matrix $\mathbf{S}$. From the Krogager decomposition, we can deduce the vector $\displaystyle \mathbf{h} = \left[ k_{d} \;\; k_{h} \;\; k_{s} \right]^T$, which is usually used to plot an RGB image.
\subsubsection{Cameron Decomposition}
The Cameron decomposition is a coherent target decomposition that exploits monostatic reciprocity (which forces the off-diagonal terms of the Sinclair matrix to be equal) and mirror symmetry (the existence of a symmetry axis perpendicular to the radar line of sight) to derive a compact complex descriptor of scattering \cite{cameron1990feature, cameron2002simulated, cameron2006conservative}. Starting from the Pauli scattering vector defined in Section \ref{sec:pauli}, the anti-symmetric component vanishes under reciprocity, and one represents the normalized complex parameter as:
\begin{equation}
     z \;=\;\frac{k_{2} + j\,k_{3}}{k_{1}}
\end{equation}
In the complex $z$–plane, eleven prototype values \(z_p\) are assigned to various scattering responses. Each resolution cell is then classified by:
\begin{equation}
    \hat p \;=\;\arg\min_{p}\,\lvert z - z_{p}\rvert
\end{equation}
As such, both the amplitude and the dominant coherent mechanism are estimated per pixel. A classification procedure can be defined based on the $z$–plane.
\subsection{Non-Coherent Polarimetric Decomposition}
\begin{figure}[htbp]
    \centering
    \includegraphics[width=1\columnwidth]{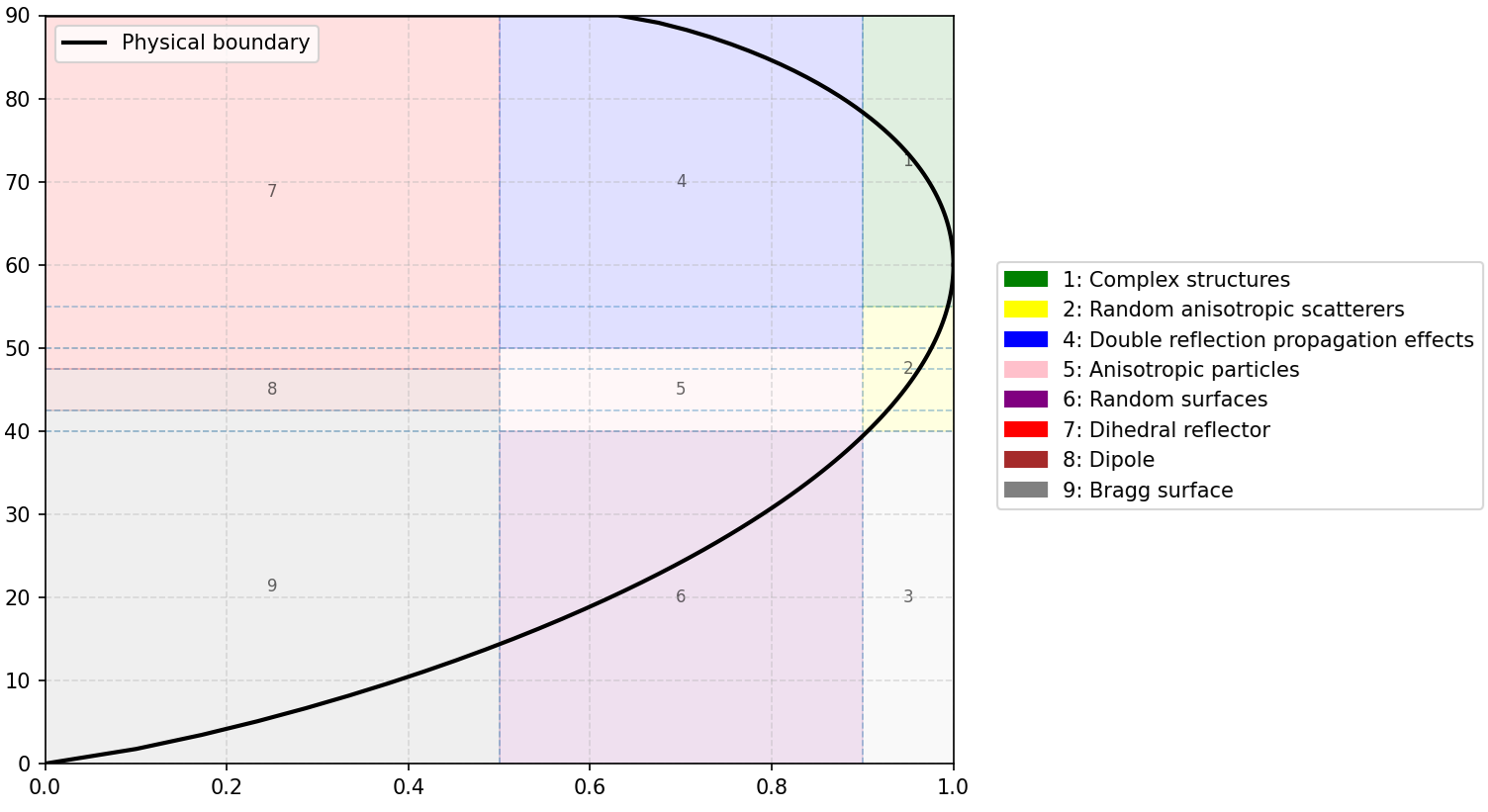}
    \caption{$H-\alpha$ plane separated into areas 1 to 9, each corresponding to a specific scattering mechanism, with the entropy at the x-axis and the scattering angle at the y-axis. The black line represents the boundary of physically possible $H-\alpha$ couples.}
    \label{fig:Halpha}
\end{figure}

Although coherent decompositions help describe artificial targets, they are not suited to analyze random scattering effects (forests, fields, vegetation, etc.). To fill this lack, the Pauli vector $\mathbf{k}$ is usually modeled by the multivariate, centered, circular complex Gaussian distribution $\mathcal{CN}(\mathbf{0}, \mathbf{T})$ which is fully characterized by the covariance matrix $\mathbf{T}=E\left[\mathbf{k} \,\mathbf{k}^H\right]$. This covariance matrix is estimated locally on a PolSAR image through the Sample Covariance Matrix (SCM) $\hat{\mathbf{T}}$ computed on a set of $N$ samples taken in a spatial boxcar: 
\begin{equation}
    \hat{\mathbf{T}} = \frac{1}{N} \sum_{i=1}^N \mathbf{k}_i\, \mathbf{k}_i^H\, . 
    \label{eq:coherence}
\end{equation}
The entropy $H$ indicates the randomness of the overall backscattering phenomenon: $H = \displaystyle\sum_{i=1}^3  p_i \, \log{p_i}$ with the pseudo-probabilities  $p_i = \lambda_i /\left(\sum_{j=1}^3 \lambda_j \right) $ where $\lambda_i$ are the eigenvalues of the SCM $\hat{\mathbf{T}}$. Entropy $H$ varies between $0$ and $1$. A low entropy indicates that the observed target is pure and the backscattering is deterministic. This is reflected by a single non-zero normalized eigenvalue close to $1$. When the entropy is high, it reflects the completely random nature of the observed target. This occurs when the pseudo-probabilities are identical.
The angle $\alpha$ is defined as $\alpha =  \displaystyle\sum_{i=1}^{3} p_i\, \alpha_i$ where $\alpha_i = \arccos(|e_{i1}|)$ and where $e_{i1}$ is the first component of the $i$-th eigenvector of the SCM. It varies between $0^\circ$ and $90^\circ$ and characterizes the type of the dominant scattering mechanism (surface diffusion, dihedral diffusion). The relationship between entropy, $\alpha$ angle, and scattering mechanisms is represented in Figure~\ref{fig:Halpha}.
A classification procedure can be defined based on the entropy and alpha parameters. Indeed, by considering the two-dimensional $H-\alpha$ space, all random scattering mechanisms can be represented \cite{formont2010statistical}. Therefore, a pixel belonging to a region of the $H-\alpha$ plane allows a physical interpretation of the average scattering mechanism.
\section{Datasets}
\label{app:datasets}
\subsection{San Francisco Polarimetric SAR ALOS-2} 
The San Francisco Polarimetric SAR ALOS-2 dataset\footnote{https://ietr-lab.univ-rennes1.fr/polsarpro-bio/san-francisco/} is an open-source PolSAR dataset. More precisely, it is an L-band polarimetric SAR image with a ground range resolution approximating $10$m that has been acquired by the satellite ALOS-2: due to the penetration capability of the L-band wave into forest, vegetation, snow, and soil medium, ALOS-2 brought precious information on the earth surface objects  \cite{yamaguchi2016alos}. Note that the San Francisco Polarimetric SAR ALOS-2 dataset is a \textit{full polarimetric SAR} data, i.e., the four channels of the image represent the four elements of the Sinclair matrix. Due to the monostatic SAR context, we transform it into a three-channel image.
The dataset is built by cropping the $22,608 \times 8080$ image into $64\times 64$ non-overlapping tiles, resulting in a dataset of $44,478$ three-channel tiles. Tiles are randomly assigned to the training, validation, and test folds with $70\%$ for training, $15\%$ for validation, and $15\%$ for test.
Note that the image in the experiments is a crop of the original image to reduce computation cost, resulting in a $4200 \times 2000$ image.
\subsection{PolSF}
\label{app:setup_polsf}
The PolSF dataset is a segmentation dataset segmentation mask \cite{liu2022polsf}. The segmentation mask comprises $6$ classes (plus one unlabeled class) corresponding to various terrain types (e.g., water, forests, urban areas). The dataset is strongly imbalanced as some classes are more represented than others.
Similar to the San Francisco Polarimetric SAR ALOS-2 dataset, the PolSF dataset is constructed by cropping the $4200 \times 2000$ image into $64 \times 64$ non-overlapping tiles, resulting in a dataset of $3397$ three-channel tiles. Tiles are randomly assigned with a sampling weight based on their majority class to the training, validation, and test folds, with $70\%$ allocated to training, $15\%$ to validation, and $15\%$ to test.
\subsection{S1SLC\_CVDL}
\label{app:setup_s1slc}
The \text{S1SLC\_CVDL}\footnote{https://ieee-dataport.org/open-access/s1slccvdl-complex-valued-annotated-single-look-complex-sentinel-1-sar-dataset-complex} is an open-source PolSAR dataset \cite{nm4g-yd98-23}. It is a C-band polarimetric SAR image with a ground range resolution  \cite{geudtner2014sentinel}. This dataset comprises $276,571$ two-channel images semantically annotated in 7 different classes. Note that the \text{S1SLC\char`_CVDL} dataset is a \textit{dual-polarimetric SAR} dataset, i.e., the two channels of the images represent the $S_{HH}$ and $S_{HV}$ elements of the Sinclair matrix.
Like the PolSF dataset, tiles are randomly assigned with a sampling weight based on their majority class to the training, validation, and test folds with $70\%$ allocated to training, $15\%$ to validation, and $15\%$ to test.
\begin{figure}[htbp]
    \centering
    \includegraphics[width=0.9\columnwidth]{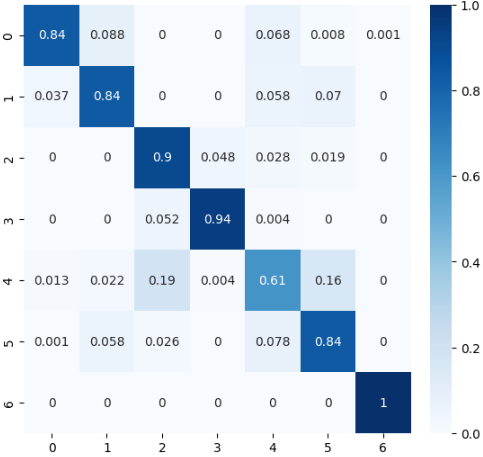}
    \caption{Confusion matrix of a CVNN LPS $\mathrm{MLP}$ for classification on \text{S1SLC\char`_CVDL}}
    \label{fig:mlp_s1slc}
\end{figure}
\begin{table}[htbp]
\centering
\resizebox{\linewidth}{!}{
\begin{tabular}{|c|c|c|c|c|c|c|c|c|c|c|c|}
\hline
Model & Trainable Parameters & Inference Time (ms) \\
\hline
ResNet LPS MLP & $2,447,560$ & $48.82\pm0.08$ \\
\hline
ResNet LPS PolyDec & $2,446,100$ & $38.56\pm0.17$ \\
\hline
ResNet LPS Norm & $2,445,995$ & $18.70\pm0.07$ \\
\hline
ResNet APS & $1,226,467$ & $7.56\pm0.30$ \\
\hline
\end{tabular}
}
\caption{Trainable parameters and inference time for shift-equivariant CVNN variants on the \text{S1SLC\_CVDL} dataset. Inference time is reported on $50$ iterations. CVNN Learnable Polyphase Sampling (LPS) variants are more costly than the shift-equivariant Adaptive Polyphase Sampling (APS) model, both in number of trainable parameters and in inference time.}
\label{tab:s1slc_stats}
\end{table}
\section{Experimental setup}
\label{app:setup}
\subsection{Classification}
\label{app:setup_classif}
For the training of the ResNet, we have used the AdamW optimizer \cite{loshchilov2017decoupled} with a weight decay of $10^{-5}$ and a learning rate varying between $10^{-2}$ and $10^{-3}$ depending on the projection function. Indeed, memory-intensive projection functions like $\mathrm{PolyDec}$ and $\mathrm{MLP}$ necessitated reducing the batch size and, by extension, the learning rate. The model is trained for $100$ epochs. For both RVNNs and CVNNs, we set the number of layers in the encoder to $4$, with an initial number of channels set to $16$. To address the class imbalance issue, we have implemented a complex-valued version of Focal Loss \cite{lin2017focal} for CVNNs and utilized the real-valued version of Focal Loss for RVNNs. Finally, we have used the $\mathrm{ReLU}$ activation for RVNNs, and $\mathrm{modReLU}$ for CVNNs. For the shift-equivariant models relying on the Gumbel Softmax, we have set the initial temperature value  $10^{-5}$. Table~\ref{tab:s1slc_stats} shows that the shift-equivariant CVNN LPS variants have a similar amount of trainable parameters, but differ greatly during inference time. We see a clear correlation between the number of trainable parameters and the inference time: as such, the $\mathrm{Norm}$ variant is the least costly in time and memory, while the $\mathrm{MLP}$ variant is the most demanding one. Interestingly, we notice that the projection layer does not have a great impact on the number of trainable parameters (switching from LPS to APS has a far greater impact on this aspect). 
\subsection{Semantic segmentation}
\label{app:setup_seg}
\begin{table}[htbp]
\centering
\resizebox{\linewidth}{!}{
\begin{tabular}{|c|c|c|c|c|c|c|c|c|c|c|c|}
\hline
Model & Trainable Parameters & Inference Time (ms) \\
\hline
UNet LPS MLP & $3,297,660$ & $29.20\pm0.21$ \\
\hline
UNet LPS PolyDec & $3,296,200$ & $23.05\pm0.07$ \\
\hline
UNet LPS Norm & $3,296,095$ & $14.10\pm0.27$ \\
\hline
UNet APS & $2,075,588$ & $11.75\pm0.40$ \\
\hline
\end{tabular}
}
\caption{Trainable parameters and inference time for shift-equivariant CVNN variants on the PolSF dataset. Inference time is reported on $50$ iterations. CVNN Learnable Polyphase Sampling (LPS) variants are more costly than the shift-equivariant Adaptive Polyphase Sampling (APS) model, both in number of trainable parameters and in inference time.}
\label{tab:polsf_stats}
\end{table}
For the training of the UNet, we have chosen the AdamW optimizer \cite{loshchilov2017decoupled} with a weight decay of $5 \times 10^{-4}$ and a learning rate of $10^{-3}$. The model is trained for $500$ epochs. For both RVNNs and CVNNs, we set the number of layers in the encoder and the decoder to $4$, with an initial number of channels set to $16$. The rest of the setup is similar to the procedure presented in Appendix \ref{app:setup_classif}. Table~\ref{tab:polsf_stats} shows that the shift-equivariant CVNN LPS variants have a similar amount of trainable parameters, but differ greatly during inference time. We see a clear correlation between the number of trainable parameters and the inference time: as such, the $\mathrm{Norm}$ variant is the least costly in time and memory, while the $\mathrm{MLP}$ variant is the most demanding one. Interestingly, we notice that the projection layer does not have a great impact on the number of trainable parameters (switching from LPS to APS has a far greater impact on this aspect).
\begin{figure}[htbp]
    \centering
    \includegraphics[width=0.9\columnwidth]{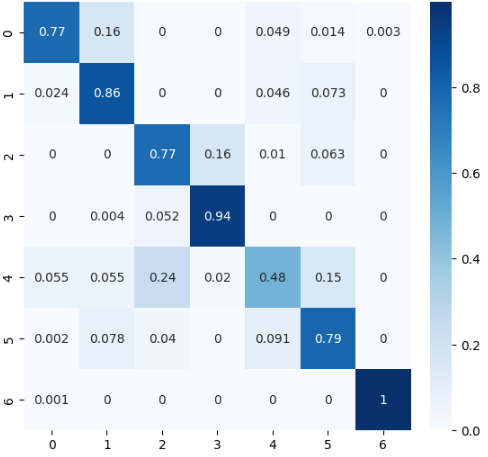}
    \caption{Confusion matrix of a CVNN LPF for classification on \text{S1SLC\char`_CVDL}}
    \label{fig:lpf_s1slc}
\end{figure}
\subsection{Reconstruction}
\label{app:setup_recon}
\begin{table}[htbp]
\centering
\resizebox{\linewidth}{!}{
\begin{tabular}{|c|c|c|c|c|c|c|c|c|c|c|c|}
\hline
Model & Trainable Parameters & Inference Time (ms) \\
\hline
AE LPS MLP & $2,898,626$ & $88.98\pm0.03$ \\
\hline
AE LPS PolyDec & $2,898,042$ & $75.29\pm0.05$ \\
\hline
AE LPS Norm & $2,898,000$ & $46.50\pm0.02$ \\
\hline
AE APS & $1,750,348$ & $24.23\pm0.01$ \\
\hline
\end{tabular}
}
\caption{Trainable parameters and inference time for shift-equivariant CVNN variants on the San Francisco Polarimetric SAR ALOS-2 dataset. Inference time is reported on $50$ iterations. CVNN Learnable Polyphase Sampling (LPS) variants are more costly than the shift-equivariant Adaptive Polyphase Sampling (APS) model, both in number of trainable parameters and in inference time.}
\label{tab:alos2_stats}
\end{table}
For the training of the AutoEncoder, we have chosen the AdamW optimizer \cite{loshchilov2017decoupled} with a weight decay of $0$ and a learning rate of $5\times10^{-4}$. The model is trained for $50$ epochs. We set the number of layers in the encoder and the decoder to $2$, with an initial number of channels to $64$. As expected from the unsupervised nature of the reconstruction task, we have used the Mean Squared Error loss function. Finally, we have used the $\mathrm{ReLU}$ activation for RVNNs, and $\mathrm{modReLU}$ for CVNNs. For the shift-equivariant models relying on the Gumbel Softmax, we have set the initial temperature value to $0.001$, the gamma value to $0.1$, and the minimum value to $10^{-5}$. Table~\ref{tab:alos2_stats} shows that the shift-equivariant CVNN LPS variants have a similar amount of trainable parameters, but differ greatly during inference time. We see a clear correlation between the number of trainable parameters and the inference time: as such, the $\mathrm{Norm}$ variant is the least costly in time and memory, while the $\mathrm{MLP}$ variant is the most demanding one. Interestingly, we notice that the projection layer does not have a great impact on the number of trainable parameters (switching from LPS to APS has a far greater impact on this aspect).
\begin{figure}[htbp]
    \centering
    \includegraphics[width=\columnwidth]{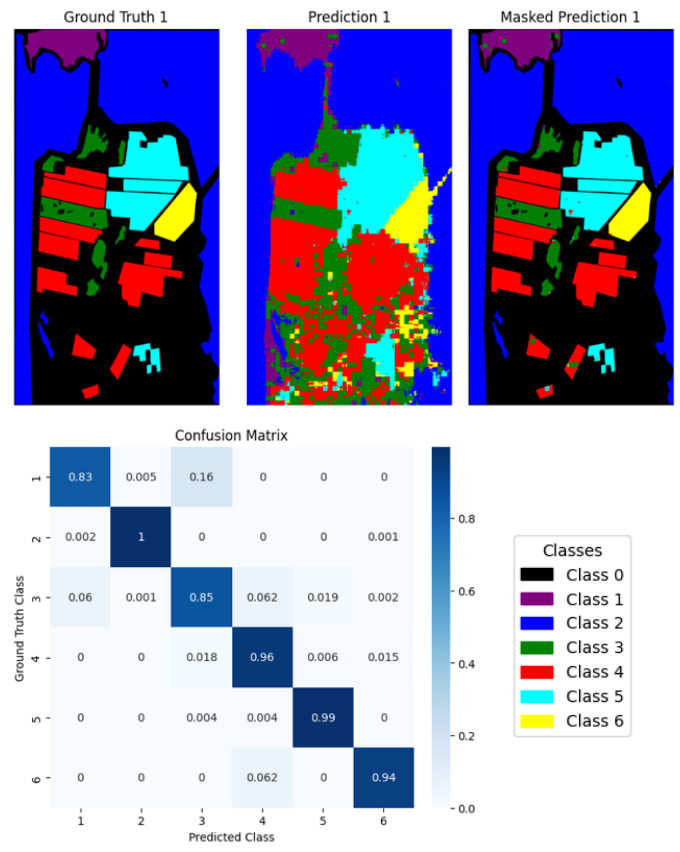}
    \caption{Results obtained with a CVNN LPS $\mathrm{PolyDec}$ for semantic segmentation on PolSF. Left: ground truth segmentation mask. Middle: the complete prediction (without a mask for the unlabeled class). Right: prediction (with a mask for the unlabeled class). Bottom: confusion matrix between ground truth and prediction}
    \label{fig:poly_polsf}
\end{figure}

\section{Additional Results}
\label{app:more_results}
\subsection{Classification}
\label{app:more_results_classif}
In addition to the results presented in Section \ref{exp:classification}, we include the confusion matrices of the CVNN LPS $\mathrm{MLP}$ and the CVNN LPF to showcase the impact of our method against a non-shift-equivariant CVNN.
As we can observe from Figures \ref{fig:mlp_s1slc} and \ref{fig:lpf_s1slc}, the confusion matrix of the  CVNN LPS $\mathrm{MLP}$ is better. The semantic classes of the \text{S1SLC\char`_CVDL} dataset are defined as follows: Agricultural fields, Forest and Woodlands, High-Density Urban Areas, High Rise Buildings, Low-Density Urban Areas, Industrial Regions, and Water Regions. The confusion matrix from Figure \ref{fig:mlp_s1slc} shows a better distinction between urban areas than the confusion matrix from Figure \ref{fig:lpf_s1slc}: High-Density Urban Areas, Low-Density Urban Areas, and Industrial Regions. This result is crucial as it highlights that our proposed approach increases the viability of neural networks for real-life applications.
\subsection{Semantic Segmentation}
\label{app:more_results_seg}
In addition to the results presented in Section \ref{exp:segmentation}, we include the visualizations of the CVNN LPS $\mathrm{PolyDec}$ and the CVNN LPF to showcase the impact of our method against a non-shift-equivariant CVNN.
\begin{figure}[htbp]
    \centering
    \includegraphics[width=\columnwidth]{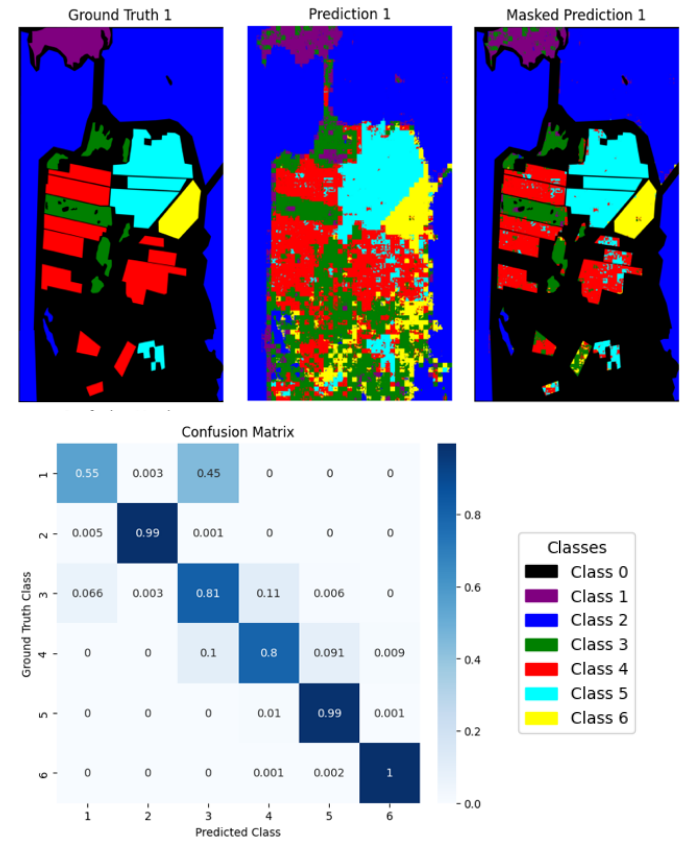}
    \caption{Results obtained with a CVNN LPF for semantic segmentation on PolSF. Left: ground truth segmentation mask. Middle: complete prediction (without a mask for the unlabeled class). Right: prediction (with a mask for the unlabeled class). Bottom: confusion matrix between ground truth and prediction}
    \label{fig:lpf_polsf}
\end{figure}
\begin{figure*}[htbp]
    \centering
    \includegraphics[width=\textwidth]{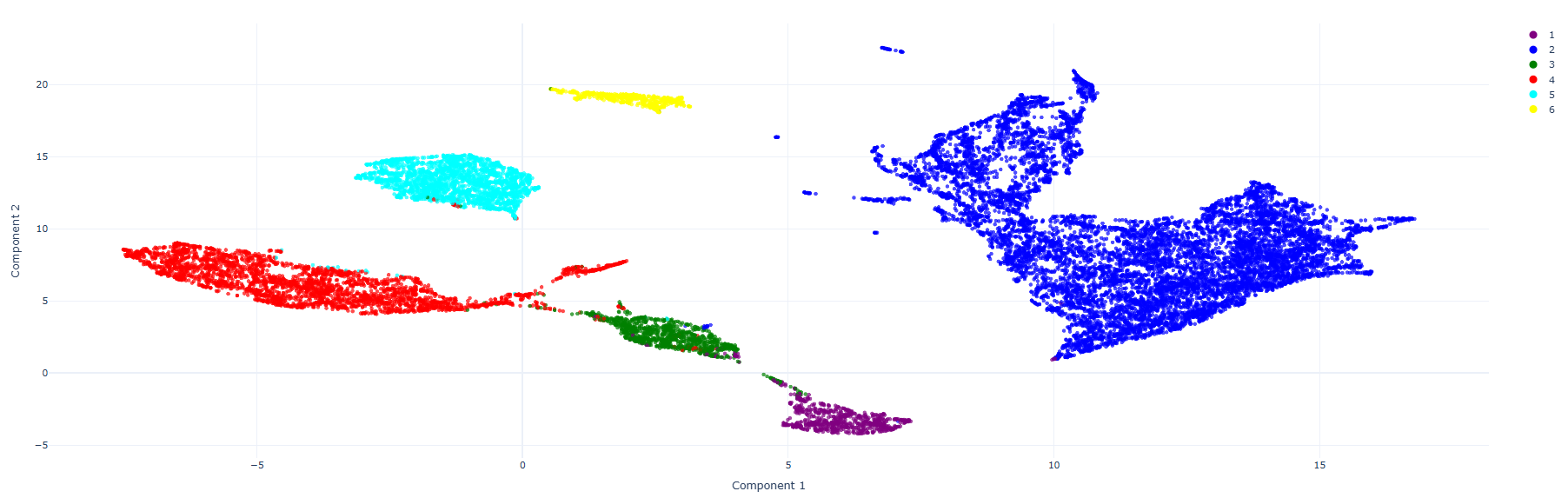}
    \caption{UMAP visualization of the latent space from CVNN LPS $\mathrm{PolyDec}$ on PolSF dataset.}
    \label{fig:umap_poly_polsf}
\end{figure*}
\begin{figure*}[htbp]
    \centering
    \includegraphics[width=\textwidth]{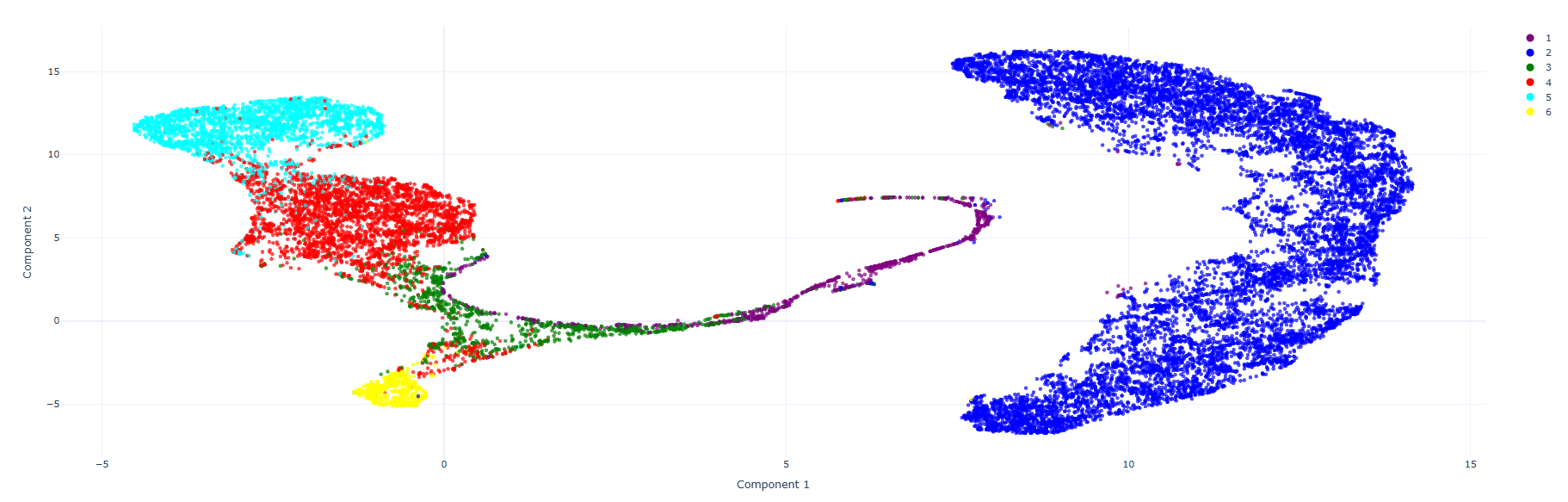}
    \caption{UMAP visualization of the latent space from CVNN LPF on PolSF dataset.}
    \label{fig:umap_lpf_polsf}
\end{figure*}
As we can observe from Figures \ref{fig:poly_polsf} and \ref{fig:lpf_polsf}, the predicted segmentation mask of the CVNN LPS $\mathrm{PolyDec}$ is smoother and closer to the ground truth. The semantic classes of the PolSF dataset are defined as follows: Mountain, Water, Vegetation, High-Density Urban, Low-Density Urban, and Developed Area. The confusion matrix from Figure \ref{fig:poly_polsf} shows a better distinction between natural areas than from Figure \ref{fig:lpf_polsf}: Mountain and Vegetation. This conclusion is further supported by latent spaces visualization from Figures \ref{fig:umap_poly_polsf} and \ref{fig:umap_lpf_polsf}. We also have fewer artifacts in the middle of correctly labeled zones (such as wrongfully predicting the presence of a forest in the middle of an urban area). As stated in \ref{app:more_results_classif}, this result highlights the advantages of our approach over traditional architectures regarding real-life applications.
\begin{figure*}[htbp]
    \centering
    \includegraphics[width=\textwidth]{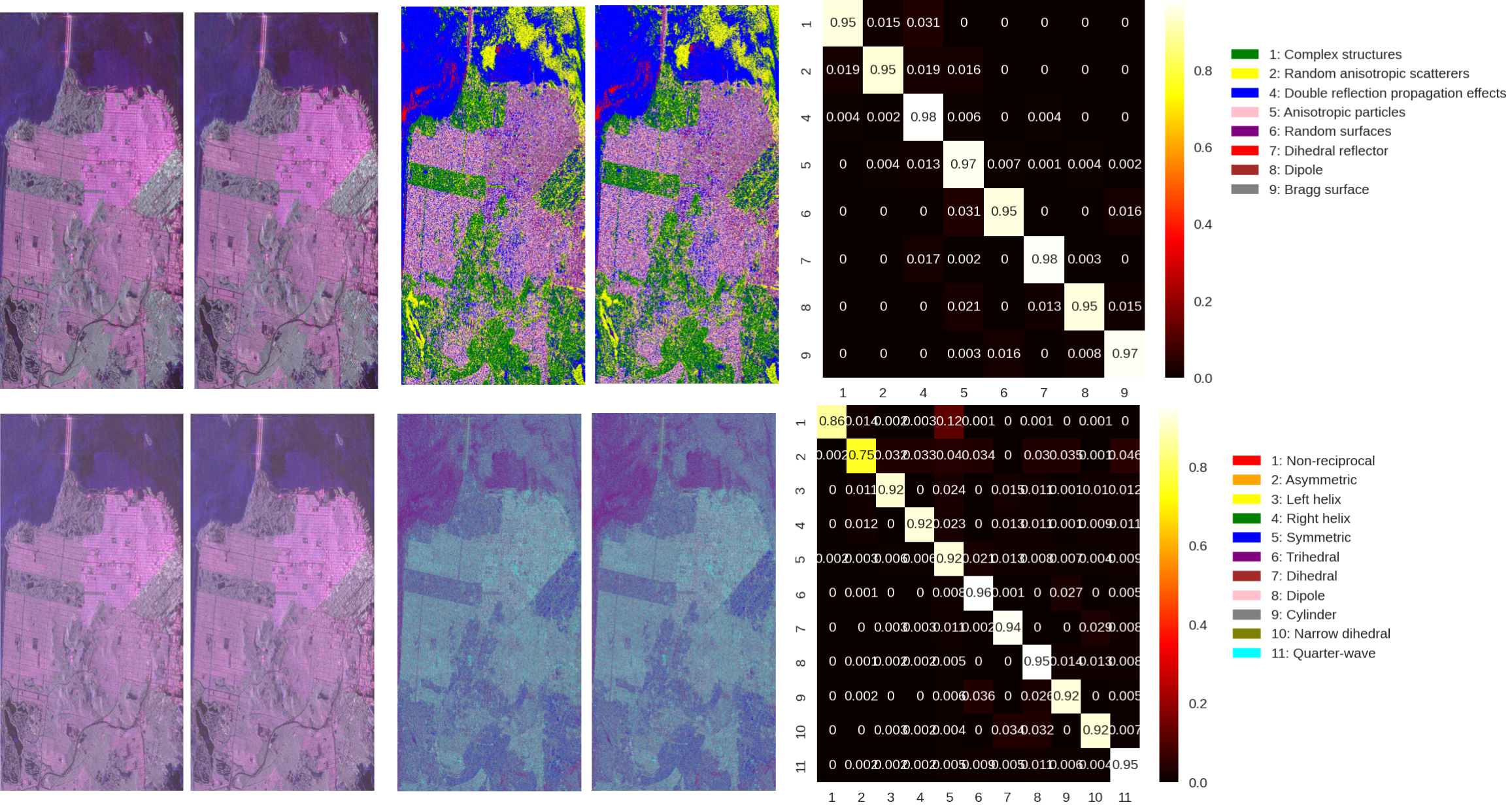}
    \caption{Reconstruction results obtained from CVNN LPS $\mathrm{PolyDec}$ on San Francisco Polarimetric SAR ALOS-2 dataset. Up-Left: Amplitude images of the original (left) and reconstructed (right) images with the Pauli basis. Up-Middle: Images of the original (left) and reconstructed (right) images with the $H-\alpha$ classification. Up-Right: Down-Right: Confusion matrix of the original (rows) and reconstructed (columns) $H-\alpha$ classes. Down-Left: Amplitude images of the original (left) and reconstructed (right) images with the Krogager basis. Down-Middle: Images of the original (left) and reconstructed (right) images with the Cameron classification. Down-Right: Confusion matrix of the original (rows) and reconstructed (columns) Cameron classes.}
    \label{fig:poly_alos2}
\end{figure*}
\begin{figure*}[htbp]
    \centering
    \includegraphics[width=\textwidth]{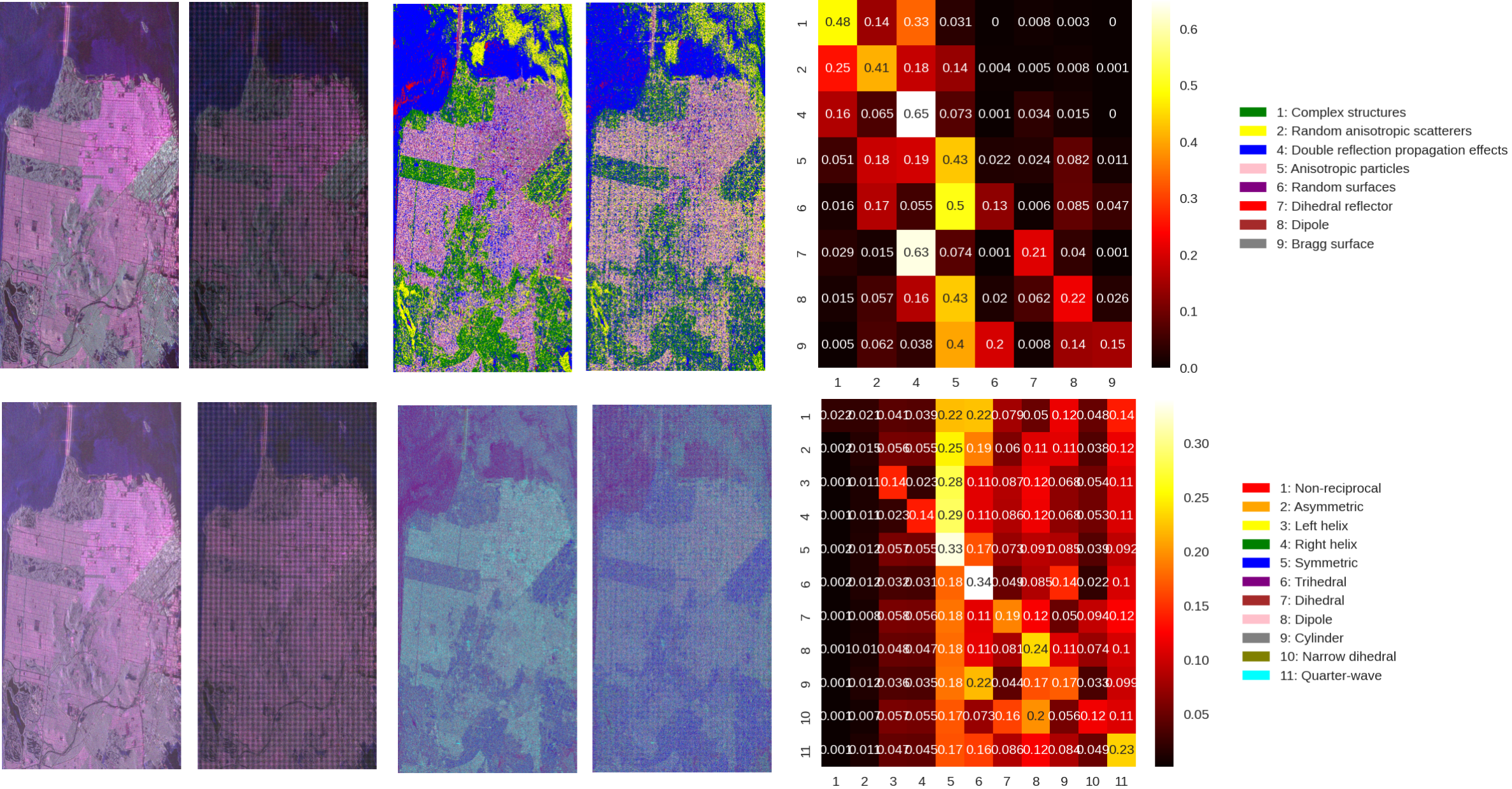}
    \caption{Reconstruction results obtained from CVNN LPF on San Francisco Polarimetric SAR ALOS-2 dataset. Up-Left: Amplitude images of the original (left) and reconstructed (right) images with the Pauli basis. Up-Middle: Images of the original (left) and reconstructed (right) images with the $H-\alpha$ classification. Up-Right: Down-Right: Confusion matrix of the original (rows) and reconstructed (columns) $H-\alpha$ classes. Down-Left: Amplitude images of the original (left) and reconstructed (right) images with the Krogager basis. Down-Middle: Images of the original (left) and reconstructed (right) images with the Cameron classification. Down-Right: Confusion matrix of the original (rows) and reconstructed (columns) Cameron classes.}
    \label{fig:lpf_alos2}
\end{figure*}
\subsection{Reconstruction}
\label{app:more_results_recon}
In addition to the results presented in Section \ref{exp:reconstruction}, we include the visualizations of the CVNN LPS $\mathrm{PolyDec}$ and the CVNN LPF to showcase the impact of our method against a non-shift-equivariant CVNN.
As we can observe from Figures \ref{fig:poly_alos2} and \ref{fig:lpf_alos2}, the reconstruction of the CVNN LPS $\mathrm{PolyDec}$ is almost perfect when compared to the ground truth.
\\
The semantic classes of the $H-\alpha$ are shown in Figure \ref{fig:Halpha} and defined as follows: Complex structures, Random anisotropic scatterers, Double reflection propagation effects, Anisotropic particles, Random surface, Dihedral reflector, Dipole, and Bragg surface. 
Similarly, the semantic classes of the Cameron are defined as follows: Non-reciprocal, Asymmetric, Left helix, Right helix, Symmetric, Trihedral, Dihedral, Dipole, Cylinder, Narrow dihedral, and Quarter-wave.
\\
The various visualizations allow us to make a general observation: polarimetric decompositions (Pauli, Krogager, and $H-\alpha$) and reconstruction metrics (amplitude and angular distances) are incredibly better for the CVNN LPS $\mathrm{PolyDec}$ model \ref{fig:poly_alos2} when compared to the CVNN LPF model \ref{fig:lpf_alos2}. We believe that such results show promising perspectives regarding further experiments on the impact of CVNNs on the reconstruction of PolSAR images.

\end{document}